\documentclass[]{TEAI}
\usepackage{helvet}

\usepackage{amsmath} 
\usepackage{natbib}
\usepackage{graphicx}
\usepackage{subcaption} 

\usepackage[toc,page,header]{appendix}
\usepackage[utf8]{inputenc} % allow utf-8 input
\usepackage[T1]{fontenc}    % use 8-bit T1 fonts
\usepackage{hyperref}       % hyperlinks
\usepackage{url}            % simple URL typesetting
\usepackage{booktabs}       % professional-quality tables
\usepackage{lmodern}        % scalable Latin Modern fonts
\usepackage{amsfonts}       % blackboard math symbols
\usepackage{nicefrac}       % compact symbols for 1/2, etc.
\usepackage{microtype}      % microtypography
\usepackage{wrapfig}

\usepackage{amssymb}  % 用于特殊符号如♠♣等
\usepackage{fontawesome}  % 用于邮箱图标 \faEnvelope
\usepackage{url}  % 用于 \url 命令

\usepackage{titletoc}

\usepackage{tikz}  % 用于绘制彩色标记符号
\usepackage{comment}  % 用于注释备用版本
\usepackage{tabularx}  % 如果需要自动调整列宽
\usepackage{booktabs}  % 用于更美观的表格线条(可选)
\usepackage{minitoc}

\usepackage{booktabs}
\usepackage{array}
\usepackage{etoolbox}

\definecolor{lightblue}{RGB}{200, 230, 255}  
\definecolor{headerblue}{RGB}{150, 200, 255} 

\usepackage{pgfplots}
\usepackage[utf8]{inputenc} % allow utf-8 input
\usepackage[T1]{fontenc}    % use 8-bit T1 fonts
\usepackage{hyperref}       % hyperlinks
\usepackage{url}            % simple URL typesetting
\usepackage{booktabs}       % professional-quality tables
\usepackage{amsfonts}       % blackboard math symbols
\usepackage{nicefrac}       % compact symbols for 1/2, etc.
\usepackage{microtype}      % microtypography
\usepackage{xcolor}         % colors
\usepackage{graphicx}
\usepackage{float}
\usepackage{comment}
\usepackage{multirow} % For multi-row cells
\usepackage{amsmath} % For \text command if needed inside math mode\Delta
\usepackage{makecell} % For multi-line cells and better vertical spacing in cells
\usepackage{siunitx}  % For better number alignment (optional but recommended)
\usepackage{tikz}
\usepackage{pgf-pie} % Package for creating pie charts
\usepackage{subcaption}
\usepackage{wrapfig}
\usepackage[export]{adjustbox}

\usepackage{ragged2e}      % for \RaggedRight in tabularx
\usepackage{tabularx}       % For tables with fixed total width and auto-adjusting columns
\usepackage{array}          % For advanced column formatting (like >{\centering\arraybackslash}X)
\usepackage{caption}        % Recommended for figures/tables, but we'll do simple text below images here.
\usepackage{enumitem}
\usepackage{pifont}
\usepackage[hang,flushmargin]{footmisc} % 更好的脚注处理

\usepackage{tcolorbox}
\usepackage{tcolorbox}    % 1. 先加载 tcolorbox 主包
\tcbuselibrary{breakable}  % 2. 显式加载 breakable 库
\tcbuselibrary{skins}      % 3. 显式加载 skins 库 (这个库提供 topruleatbreak 等选项)
\usepackage{tabularx}
\usepackage{listings}
\usepackage{xspace}
\pgfplotsset{compat=1.18}
\usepackage{xspace}
\newcommand{\slowgen}{\textit{SlowGenerator}\xspace}
\newcommand{\slowadapt}{\textit{SlowAdapter}\xspace}
\newcommand{\fastgen}{\textit{FastGenerator}\xspace}
\newcommand{\fastadapt}{\textit{FastAdapter}\xspace}

\title{\hspace{-5pt}\includegraphics[height=1.5em]{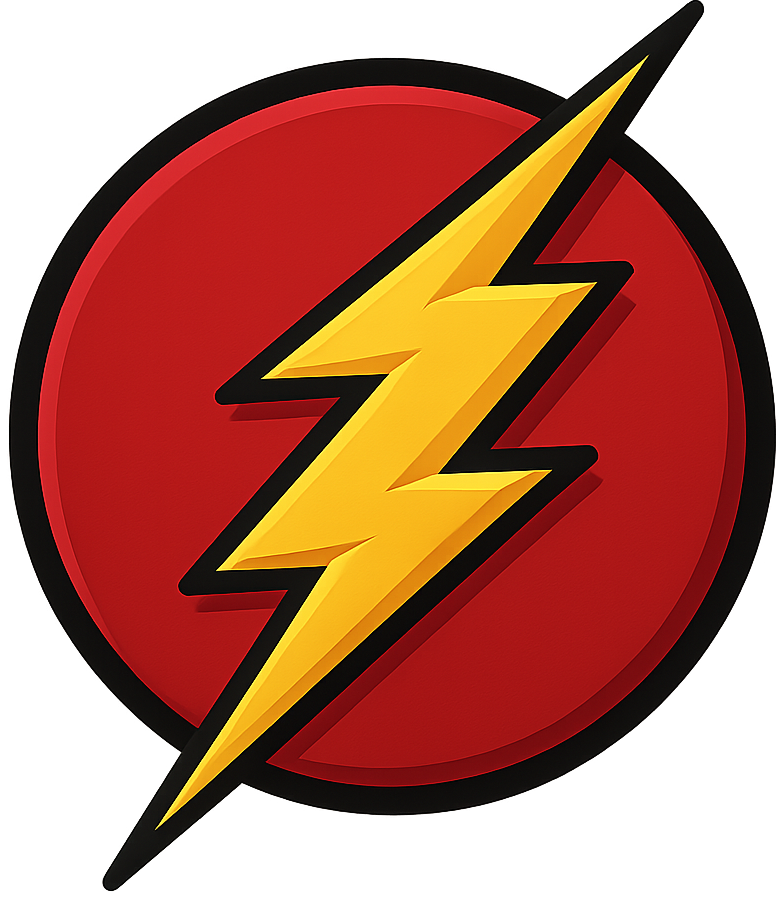}
\textsc{FlashMotion: Few-Step Controllable Video Generation with Trajectory Guidance}}

\author{
Quanhao Li\textsuperscript{1,2} \quad
Zhen Xing\textsuperscript{1,2} \quad
Rui Wang\textsuperscript{1,2} \quad
Haidong Cao\textsuperscript{1,2}\\
Qi Dai\textsuperscript{3} \quad
Daoguo Dong\textsuperscript{1,2} \quad
Zuxuan Wu\textsuperscript{1,2,$\dagger$}
}

\affiliation[1]{\mbox{Institute of Trustworthy Embodied AI, Fudan University}} 
\affiliation[2]{\mbox{Shanghai Key Laboratory of Multimodal Embodied AI}} 
\affiliation[3]{\mbox{Microsoft Research Asia}}

\abstract{
\begin{abstract}

Recent advances in trajectory-controllable video generation have achieved remarkable progress. Previous methods mainly use adapter-based architectures for precise motion control along predefined trajectories.
However, all these methods rely on a multi-step denoising process, leading to substantial time redundancy and computational overhead.
While existing video distillation methods successfully distill multi-step generators into few-step, directly applying these approaches to trajectory-controllable video generation results in noticeable degradation in both video quality and trajectory accuracy.
To bridge this gap, we introduce \textbf{FlashMotion}, a novel training framework designed for few-step trajectory-controllable video generation.
We first train a trajectory adapter on a multi-step video generator for precise trajectory control.
Then, we distill the generator into a few-step version to accelerate video generation.
Finally, we finetune the adapter using a hybrid strategy that combines diffusion and adversarial objectives, aligning it with the few-step generator to produce high-quality, trajectory-accurate videos.
For evaluation, we introduce \textbf{FlashBench}, a benchmark for long-sequence trajectory-controllable video generation that measures both video quality and trajectory accuracy across varying numbers of foreground objects. 
Experiments on two adapter architectures show that FlashMotion surpasses existing video distillation methods and previous multi-step models in both visual quality and trajectory consistency.
\end{abstract}
}

\checkdata[Code]{\url{https://github.com/quanhaol/FlashMotion}}
\checkdata[Website]{\url{https://quanhaol.github.io/flashmotion-site/}}

\begin{document}
\maketitle
\begin{figure}[h]  % 优先放在当前位置
    \centering
    \includegraphics[width=\linewidth]{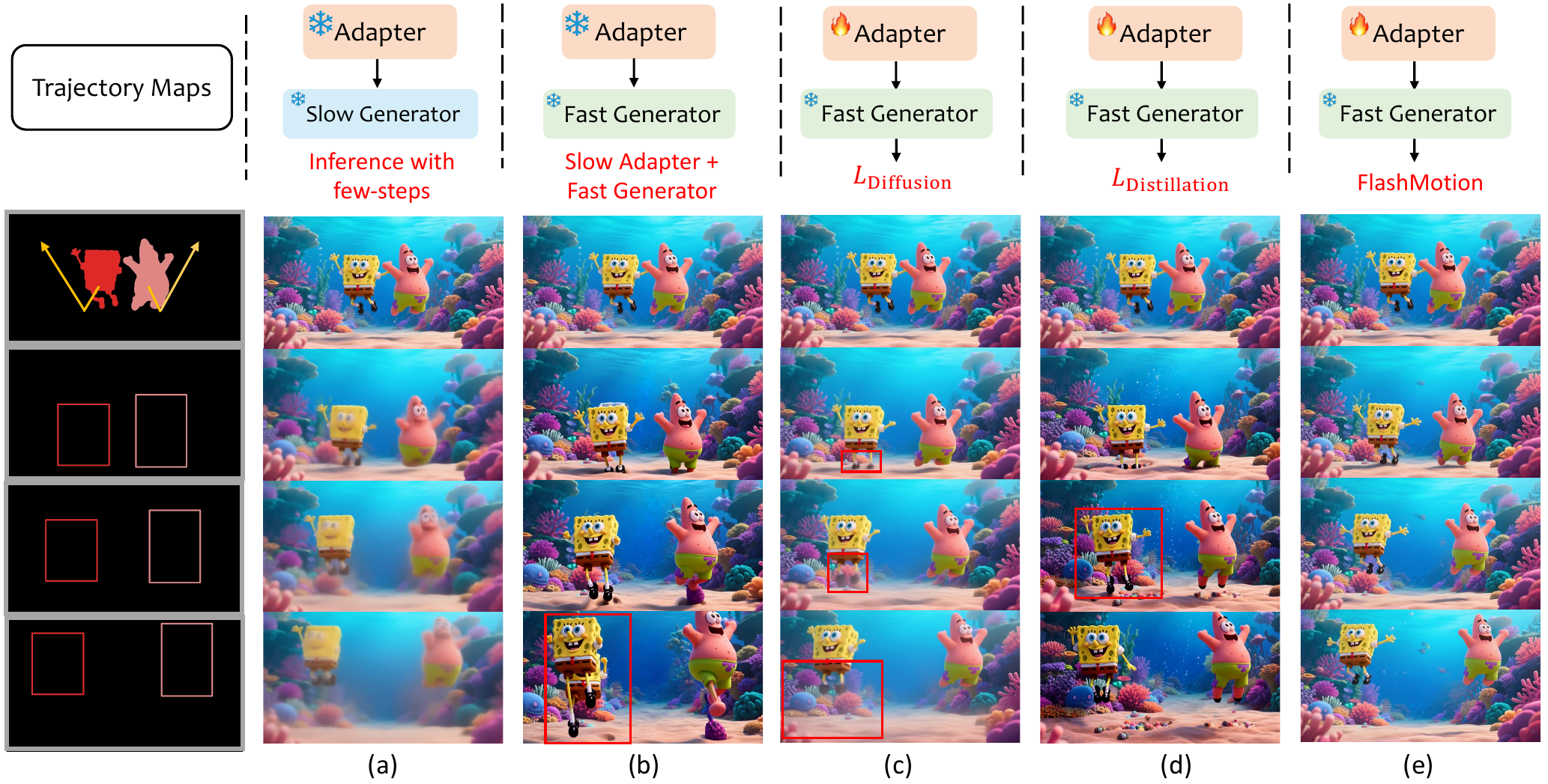}
    \caption{Illustration of the motivation and capabilities of \textbf{FlashMotion}.
    We define the \slowgen as the multi-step video model and the \fastgen as its few-step distilled version. The \slowadapt is trained with the \slowgen, while the \fastadapt is fine-tuned for the \fastgen.
    (a) Using the \slowadapt with \slowgen under few-step inference causes blurry outputs.
    (b) Applying the \slowadapt to the \fastgen degrades both quality and trajectory accuracy.
    (c) Finetuning the adapter with only diffusion loss still leads to blur artifacts.
    (d) Finetuning the adapter with existing distillation methods yields suboptimal quality and trajectory control.
    (e) FlashMotion achieves high-quality, accurate few-step trajectory-controllable video generation.}
    \label{fig:motivation}
\end{figure}
\renewcommand{\thefootnote}{}
\footnotetext{$^\dagger$Corresponding authors.}
\renewcommand{\thefootnote}{\arabic{footnote}}

\vspace{-1.5em}

\section{Introduction}
\label{sec:intro}

The emergence of diffusion models~\cite{song2020score, ho2020denoising, song2020denoising} has significantly advanced the field of video generation, enabling recent models~\cite{wan2025, yang2024cogvideox, kong2024hunyuanvideo, zheng2024open, HaCohen2024LTXVideo, xing2023vidiff, xing2024survey, xing2024aid} to synthesize high-quality videos directly from textual or visual inputs.
Building on these advances, trajectory-controllable video generation further introduces user-defined motion control, allowing videos to be generated following specified trajectory patterns~\cite{Li_2025_ICCV, wang2025levitor, zhang2025tora, zhang2025tora2, namekata2024sgi2v, geng2024motionprompting, yin2023dragnuwa}.
Despite their impressive generative capability, previous methods require multiple denoising steps, and directly using fewer steps can lead to severe blurry artifacts as shown in Fig.~\ref{fig:motivation} (a).

To address this high computational burden, recent video distillation methods have been proposed to distill multi-step teacher models into few-step student models, thereby significantly accelerating video generation process~\cite{yin2025causvid, huang2025selfforcing, lin2025diffusion, lin2025autoregressive, wang2023videolcm, cheng2025pose, shao2025magicdistillation, lv2025dcm, sun2025swiftvideo}. 
However, applying these methods directly to trajectory-controllable video generation can yield suboptimal results (Fig.~\ref{fig:motivation} (d)), and the acceleration of trajectory-controllable video generation still remains largely unexplored.

One straightforward way is to directly leverage existing strategies that distill a well-trained multi-step video generator (\slowgen), such as Wan~\cite{wan2025}, CogVideoX~\cite{yang2024cogvideox}, etc, to a few-step student model (\fastgen) while leaving the original trajectory adapter (\slowadapt) unchanged. However, as shown in Fig.~\ref{fig:motivation}(b), this results in significant degradation in both video quality and trajectory accuracy, indicating that \slowadapt is not directly compatible with \fastgen. 
This incompatibility arises because \slowadapt is tailored for the multi-step denoising process of \slowgen, where trajectory conditions slowly guide the initial noise through progressive refinement.
In contrast, \fastgen synthesizes videos within only a few denoising steps, resulting in totally different denoising paths.

In this paper, we propose \textbf{FlashMotion}, a novel training framework that adapts a \slowadapt on top of a \fastgen to achieve few-step, trajectory-controllable video generation.
We observe that directly fine-tuning \slowadapt to fit \fastgen using a standard diffusion loss leads to reasonable trajectory alignment, but the generated videos suffer from strong blurring artifacts (Fig.~\ref{fig:motivation}(c)). This arises from the fact that the diffusion loss offers only pixel-level supervision without enforcing distribution-level consistency, leading to a mismatch between the generated (fake) and real data distributions.

To mitigate this issue, FlashMotion introduces a diffusion discriminator to guide the optimization of the trajectory adapter, bridging the gap between generated and real video distributions.
Specifically, we finetune the \slowadapt using a hybrid training strategy that jointly optimizes diffusion and adversarial objectives. 
The diffusion discriminator is trained to distinguish noisy real video latents from generated ones, thereby aligning their underlying data distributions.
Meanwhile, the diffusion loss provides pixel-level supervision, encouraging the model to produce trajectory-aligned videos.
To balance the two objectives and ensure stable optimization, we further introduce a dynamic diffusion loss scaling mechanism that adaptively adjusts the loss weight during training.
In addition, thanks to the strong prior provided by \slowadapt, this training stage requires only a lightweight fine-tuning of 1K steps on 4 A100 GPUs, leading to minimal training cost.

Aside from the training framework, a comprehensive benchmark is also urgently needed.
Existing benchmarks for trajectory-controllable video generation~\cite{Perazzi2016, miao2021vspw, Li_2025_ICCV} are constrained by short video durations and limited trajectory annotations. To overcome these limitations, we introduce \textbf{FlashBench}, a large-scale and comprehensive benchmark that provides trajectory annotations for long video sequences. FlashBench further groups videos into six categories based on the number of foreground objects and evaluates models in each category with respect to both visual quality and trajectory control accuracy following~\cite{Li_2025_ICCV}.
In conclusion, our main contributions are as follows:
\begin{itemize}
\item To the best of our knowledge, FlashMotion is the first work to investigate few-step trajectory-controllable video generation. We propose and systematically examine a range of potentially promising approaches, offering in-depth analysis and comparison.
\item We propose a novel three-stage training framework that integrates diffusion and adversarial objectives, enabling effective training of a trajectory adapter on top of a few-step video diffusion model. FlashMotion significantly accelerates video generation while simultaneously enhancing visual fidelity and trajectory accuracy.
\item We present FlashBench, a large-scale benchmark comprising long video sequences with detailed trajectory annotations. Extensive experiments show that FlashMotion achieves superior performance, outperforming both few-step distillation methods and multi-step trajectory-guided video generation methods.
\end{itemize}
\section{Related Works}
\label{sec:formatting}

\noindent\textbf{Trajectory Controllable Video Generation}
Trajectory-controllable video generation has recently gained considerable attention for its capability to precisely control the motion trajectories of foreground objects during the video generation process.
Some training-free methods attempt to achieve trajectory control by directly manipulating the attention map values within specific spatial regions~\cite{jain2023peekaboo, qiu2024freetraj, yang2024direct, ma2024trailblazer}.
However, due to the lack of explicit trajectory supervision, such methods often struggle to achieve consistent and temporally coherent motion control.
Recent training-based approaches introduce learnable modules for trajectory control, enabling the use of various trajectory representations as conditioning signals~\cite{Li_2025_ICCV, wang2024motionctrl, zhou2025trackgo, zhang2025tora, gu2025das, fu20243dtrajmaster, wang2025levitor, wang2025cinemaster, yariv2025through, wei2024dreamvideo, wei2024dreamvideo2, wei2025dreamrelation}.
By explicitly modeling trajectory through these structured conditions, the trajectory adapter can effectively inject fine-grained spatiotemporal control into the video generation process.
Despite their improved controllability, these methods still depend on multi-step diffusion inference with tens or even hundreds of denoising iterations, resulting in significant latency and computational cost.
In contrast, FlashMotion proposes a few-step trajectory-controllable video generation model that drastically reduces the number of denoising iterations while preserving visual quality and trajectory controllability.

\noindent\textbf{Video Diffusion Distillation}
Step distillation is a common and effective approach to accelerate diffusion models. Existing video distillation methods primarily adapt image distillation methods and can be broadly classified into three categories: consistency distillation, score distillation, and adversarial distillation.
Consistency distillation~\cite{song2023consistency, luo2023latent} enables single-step generation by directly mapping any point along the probability flow trajectory back to its origin. Methods such as VideoLCM\cite{wang2023videolcm}, T2V-Turbo\cite{li2024t2v}, and DCM\cite{lv2025dcm} extend this concept to video domain, thereby achieving efficient video synthesis with minimal sampling steps.
Score distillation~\cite{yin2024onestep, yin2024improved} focuses on minimizing the discrepancy between the score estimates of the student and teacher models. 
Recent video methods such as POSE~\cite{cheng2025pose}, MagicDistillation\cite{shao2025magicdistillation}, CausVid\cite{yin2025causvid}, and Self-Forcing\cite{huang2025selfforcing} adopt score distillation objective, aiming to approximate the same distribution of the multi-step diffusion teacher model.
Adversarial distillation\cite{goodfellow2014generative, sauer2024adversarial, sauer2024fast} instead employs a discriminator to narrow the distribution gap between real and generated samples.
In the video domain, APT\cite{lin2025diffusion} and APT2~\cite{lin2025autoregressive} leverage this strategy to perform one-step adversarial distillation, training a discriminator to distinguish real videos from those synthesized by the distilled generator.
Despite their impressive efficiency gains, existing video distillation methods are not specifically designed for trajectory-controllable video generation, often resulting in degraded visual quality and trajectory accuracy when directly applied to this task.
\section{Method}

\begin{figure}[h]
    \centering
    \includegraphics[width=\linewidth]{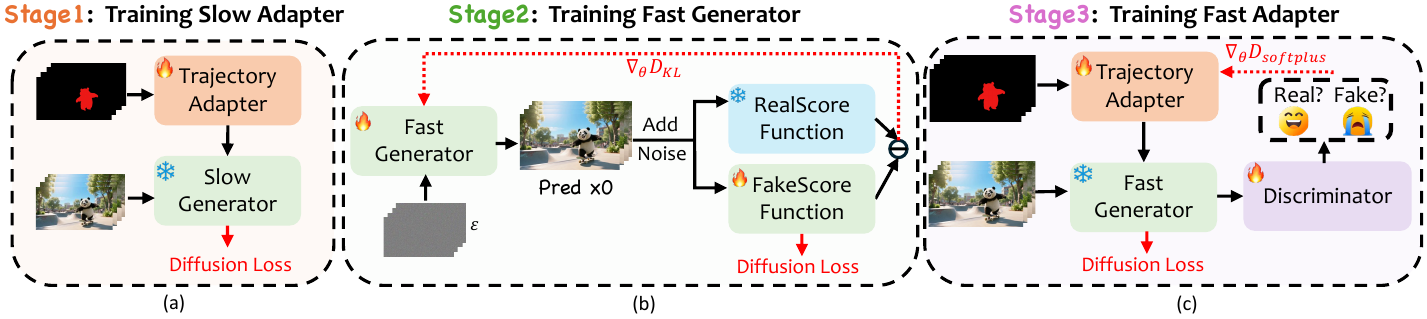}
     \caption{Overview of \textbf{FlashMotion} training pipeline. FlashMotion is trained in three stages: (1) a \slowadapt is first trained on the \slowgen with a diffusion loss; (2) a \fastgen is distilled from the \slowgen under the supervision of a distribution matching~\cite{yin2024onestep} loss; and (3) the \slowadapt is finetuned to align with the \fastgen using a hybrid training strategy that combines adversarial and diffusion losses.}
    \label{fig:training stages}
\end{figure}

\subsection{Overview}
We propose \textbf{FlashMotion}, a trajectory-controllable image-to-video framework that generates high-quality, trajectory-consistent videos in few denoising steps, achieving both controllability and efficiency.
As illustrated in Fig.~\ref{fig:training stages}, FlashMotion achieves this goal through a three-stage training process.
In Sec.~\ref{sec:slow adapter}, we provide a detailed explanation on training \slowadapt, including its model architecture and a progressive training procedure.
In Sec.~\ref{sec:fast generator}, we detail the training of \fastgen, which is achieved by distilling a multi-step teacher model into a few-step student model.
In Sec.~\ref{sec:fast adapter}, we explain how we adapt the \slowadapt into a \fastadapt via a hybrid training scheme with both diffusion and adversarial objectives.
Finally, we introduce FlashBench in Sec.~\ref{sec:benchmark}, which is a comprehensive benchmark tailored for evaluating long-duration video sequences.

\subsection{Training Slow Adapter}
\label{sec:slow adapter}
As shown in Fig.~\ref{fig:training stages} (a), FlashMotion first trains a trajectory adapter on \slowgen with a standard diffusion loss. We next describe its architecture and training process.

\noindent\textbf{Trajectory Adapter Architecture}
We design two distinct trajectory adapter architectures to evaluate the generalization ability of FlashMotion: a ControlNet-based adapter~\cite{zhang2023adding} and a lightweight ResNet-based adapter~\cite{He_2016_CVPR}. Specifically, the number of blocks in our Trajectory Adapter is kept identical to that of the DiT~\cite{Peebles2022DiT} blocks in Wan2.2-TI2V-5B~\cite{wan2025}.
A pretrained 3D VAE~\cite{kingma2013auto} encoder is used to encode the trajectory maps into a latent space $Z_{trajectory} \in R^{\frac{T}{4} \times \frac{H}{16} \times \frac{W}{16} \times 48}$, which later serves as input to our Trajectory Adapter.
The output from each Trajectory Adapter block is then passed through a zero-initialized convolution layer and added to the corresponding DiT block in the base model~\cite{zhang2023adding, xing2024simda}, thereby providing trajectory guidance.

\noindent\textbf{Training Procedure}
Following MagicMotion~\cite{Li_2025_ICCV}, we adopt a dense-to-sparse training strategy to progressively enhance the adapter’s trajectory understanding. The adapter is first trained with segmentation masks as dense trajectory conditions, and subsequently finetuned with bounding boxes as sparse trajectory conditions. Through this two-stage training process, we obtain the \slowadapt which can provide trajectory guidance to \slowgen.

\subsection{Training Fast Generator}
\label{sec:fast generator}
We aim to distill \slowgen into a \fastgen that can generate high quality video sequences within only a few denoising steps.
Specifically, we adopt Wan2.2-TI2V-5B~\cite{wan2025} as our \slowgen, which is built upon the DiT~\cite{Peebles2022DiT} architecture and employs a stack of transformer~\cite{vaswani2017attention} blocks for iterative denoising.

For distillation, we employ DMD~\cite{yin2024onestep}, a score distillation method that aligns the teacher and student video distributions $p_{\text{real}}$ and $p_{\text{fake}}$ by minimizing their Kullback–Leibler (KL) divergence.
We here consider three components: a few-step student generator $G_\theta$, a real score model $\mu_{\text{real}}$, and a fake score model $\mu_{\text{fake}}$, all initialized from the weights of Wan2.2-TI2V-5B~\cite{wan2025}.
As shown in Fig.~\ref{fig:training stages}(b), We first perform a few-step inference process with $G_\theta$ which maps pure Gaussian noise $\epsilon\sim\mathcal{N}(0,I)$ to clean video samples $x_0$. These clean samples are subsequently perturbed with additive Gaussian noise of varying magnitudes to produce diffused videos $x_t$.
These perturbed samples are then passed to the real score model $\mu_{\text{real}}$ and the fake score model $\mu_{\text{fake}}$, which respectively estimate the scores of the real and generated video distributions, defined as $s_{\text{real}}(x_t, t) = \nabla_x\log{p_{\text{real}}(x_t, t)} $, $s_{\text{fake}}(x_t, t) = \nabla_x\log{p_{\text{fake}}(x_t, t)}$.

Finally, our student generator model $G_\theta$ can be updated by the following distribution matching gradient: 
\begin{equation}
\begin{aligned}
\nabla \mathcal{L}_{\mathrm{DMD}}
&=\mathbb{E}_{t}\left(\nabla_{\theta} \operatorname{KL}\left(p_{\text {fake}} \| p_{\text {real}}\right)\right) \\
&=\underset{\epsilon\sim N(0;I)}{\mathbb{E}}\left[-\left(s_{\text {real }}\left(x_{t}, t\right)-s_{\text {fake }}\left(x_{t}, t\right)\right) \frac{d G_\theta}{d \theta}\right]
\end{aligned}
\end{equation}

During training, we freeze the real score model $\mu_{\text{real}}$ as the target distribution. Besides, we dynamically update the fake score model $\mu_{\text{fake}}$ by minimizing a standard diffusion loss, to track the evolving sample distribution produced by the student generator $G_\theta$. 
\begin{equation}
\mathcal{L}_{\mathrm{\text{fake}}}=\mathbb{E}\left[\left \| \mu_{\text{fake}}(x_t, t) - x_0 \right \| _{2}^{2}\right]
\end{equation}
where $x_0$ denotes the fake video samples generated by $G_\theta$.

\subsection{Training Fast Adapter}
\label{sec:fast adapter}

\begin{figure}[t]
    \centering
    \includegraphics[width=\linewidth]{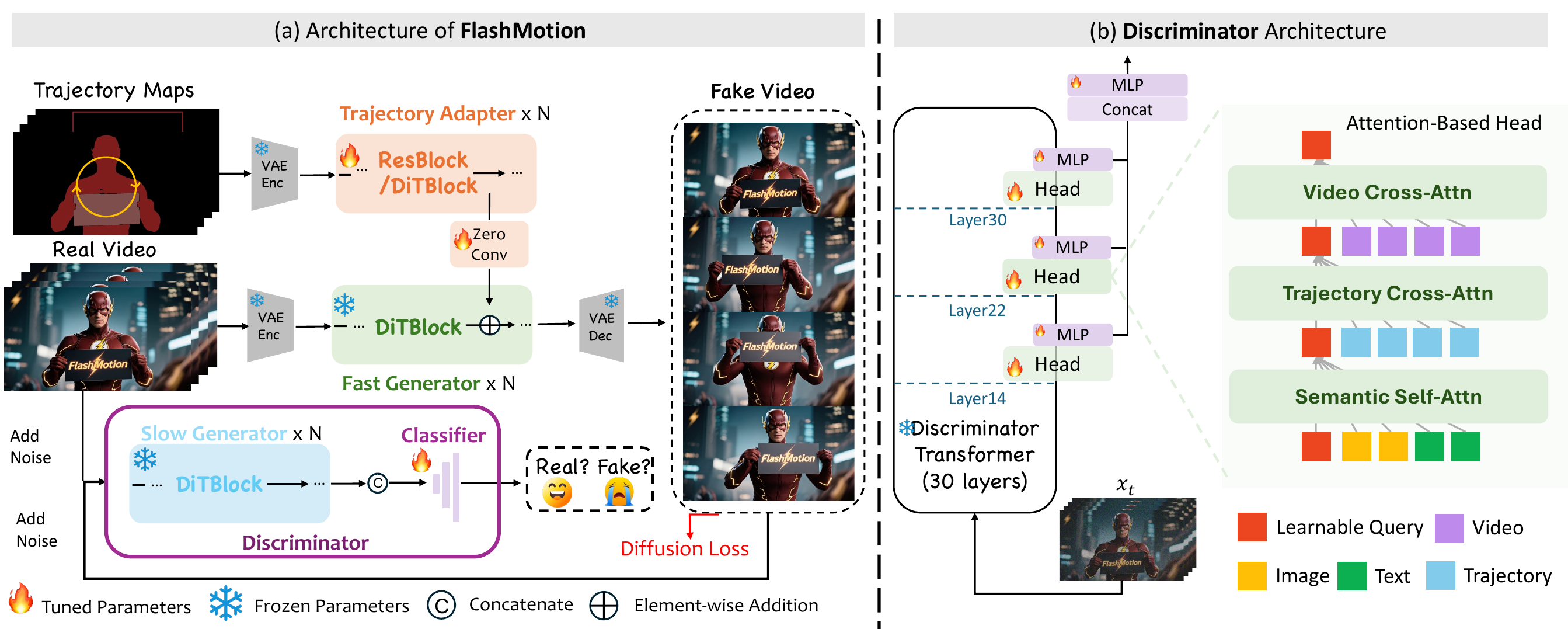}
    \caption{
    (a) Architecture of \textbf{FlashMotion}.
    The trajectory adapter is finetuned upon the \fastgen with a hybrid strategy that combines both diffusion and adversarial objectives.
    (b) Detailed illustration of our diffusion discriminator architecture.
    The discriminator adopts a DiT backbone cloned from the \slowgen, while several intermediate features from its DiT blocks are fed into an attention-based classifier to distinguish real videos from generated ones.
    }
    \label{fig:architecture}
    \vspace{-1em}
\end{figure}

As shown in Fig.~\ref{fig:motivation}(b), directly using the \slowadapt upon the \fastgen can lead to degraded visual quality and poor trajectory accuracy.
Thus, there is an urgent need for a simple and effective approach to fine-tune the \slowadapt into a \fastadapt. 
We adopt an hybrid training scheme that combines diffusion objectives and an adversarial objective, allowing the model to maintain trajectory accuracy and avoid visual quality degradation (Fig.~\ref{fig:training stages}(c)).

\noindent\textbf{Diffusion loss}
We begin by initializing the weights of the trajectory adapter using the parameters of \slowadapt trained in Stage 1 (see Sec.~\ref{sec:slow adapter} for details).
During training, as shown in Fig.~\ref{fig:architecture} (a), a pretrained 3D VAE encoder~\cite{wan2025} maps both the trajectory map and the real video into a latent space, denoted as $z_{traj}$ and $x_{0}^{\text{real}}$, which then serves as the input to the trajectory adapter and the video generator.
The trajectory features produced by each adapter block are injected into the corresponding block of the fast generator through a zero-initialized convolutional layer, thereby guiding the generation of the synthesized (fake) video latents $x_{0}^{\text{fake}} = G_\theta(x_t, t)$.
We then optimize the trajectory adapter using a standard diffusion loss:
\begin{equation}
\mathcal{L}_{diffusion}=\left \| G_\theta(x_t, t) -x_{0}^{\text{real}} \right \| _{2}^{2}
\end{equation}

\noindent\textbf{Adversarial Training}
However, as shown in Fig.~\ref{fig:motivation}(c), finetuning the \slowadapt solely with the diffusion loss often leads to noticeable blurry artifacts in the generated videos. Since the diffusion loss only enforces pixel-level alignment, it leads to a mismatch between the distributions of real and generated videos. To this end, we introduce a diffusion discriminator to bridge this distribution gap.

Inspired by APT~\cite{lin2025diffusion}, we use a diffused version of the real and fake video latents, denoted as $x_{t}^{\text{fake}}$ and $x_{t}^{\text{real}}$, as input to the diffusion discriminator, which is trained to produce a logit that effectively distinguishes between the real and generated (fake) videos.
We initialize the discriminator backbone using the weight of Wan2.2-TI2V-5B~\cite{wan2025}, and incorporate an attention-based classifier into the diffusion transformer to produce logits.
For memory efficiency and faster convergence, we freeze the backbone of the diffusion discriminator and only train the newly added classifier.

As shown in Fig.~\ref{fig:architecture} (b), the classifiers are attached to selected layers of the original DiT backbone.
Each classifier includes an attention-based head followed by an MLP layer that outputs a single token.
The tokens from all classifiers are then concatenated and passed through another MLP layer to produce the final logits, indicating whether the input video is real or fake.

Specifically, as illustrated in Fig.~\ref{fig:architecture} (b), each classifier block processes a learnable query token through three consecutive attention layers.
The \textit{Semantic Self-Attention} layer integrates the first-frame image and text information to enhance semantic representation.
In this layer, the learnable query token $q$ is concatenated with the first-frame image embeddings $e_i$ and text embeddings $e_{\text{text}}$, and then processed by a self-attention operation that enables the query token to attend across multiple semantic modalities.
Then, the resulting token is subsequently passed to the \textit{Trajectory Cross-Attention} layer, where it serves as the query and attends to the trajectory map tokens $e_{\text{traj}}$, used as keys and values in the attention computation~\cite{vaswani2017attention}.
Finally, the token is processed by the \textit{Video Cross-Attention} layer, attending to the video tokens $e_{\text{video}}$.
Each attention layer is followed by a residual connection applied to the learnable token, which is omitted in Fig.~\ref{fig:architecture} (b) for clarity.

We thus employ the following loss to finetune the trajectory adapter and the diffusion discriminator in an alternating scheme.
\begingroup
\setlength{\abovedisplayskip}{4pt}
\setlength{\belowdisplayskip}{4pt}
\setlength{\abovedisplayshortskip}{2pt}
\setlength{\belowdisplayshortskip}{2pt}
\begin{equation}
\mathcal{L}_{\mathcal{G}}=\min_{\theta}\mathbb{E}_{t\sim[0,T]}\!\left[f\!\left(-\mathcal{D}_\phi\!\left(x_{t}^{\text{fake}},t\right)\right)\right]
\end{equation}
\begin{equation}
\begin{aligned}
\mathcal{L}_{\mathcal{D}}
=\min_{\phi}\mathbb{E}_{t\sim[0,T]}\!\Big[
&f\!\left(-\mathcal{D}_\phi\!\left(x_{t}^{\text{real}},t\right)\right)
+ f\!\left(\mathcal{D}_\phi\!\left(x_{t}^{\text{fake}},t\right)\right)
\Big]
\end{aligned}
\end{equation}
\endgroup

where $f$ is the softplus function~\cite{NIPS2000_44968aec}, $T=1000$, $\mathcal{D}_\phi$ denotes the diffusion discriminator, $\theta$ and $\phi$ represent the parameters of the trajectory adapter and classifier.

\noindent\textbf{Dynamic Diffusion Loss Scale}
The diffusion loss enforces the generated video to follow the user-specified trajectory at the pixel level, while the GAN loss bridges the distribution gap between the generated and real videos.
Accordingly, we jointly train the trajectory adapter using a combination of these two objectives, formulated as:
\begin{equation}
\mathcal{L} = \mathcal{L}_{\mathcal{G}} + \lambda\mathcal{L}_{diffusion}
\end{equation} 
However, we observe that in the early stages of training, the gradients of the diffusion loss $\mathcal{L}_{diffusion}$ are substantially larger than those of the GAN loss $\mathcal{L}_{\mathcal{G}}$, and directly combining them can still lead to blurred results.
To mitigate this imbalance, we introduce a dynamic weighting scheme for the coefficient $\lambda$, defined as:
\begin{equation}
\lambda=\frac{1}{4} \times 10^{-3} \times step+0.1
\end{equation}
where $step$ means the current training iteration.
\subsection{FlashBench}
\label{sec:benchmark}
Previous works on trajectory-controllable video generation~\cite{zhang2025tora, li2025image, shi2024motion, objctrl2.5d, zhou2025trackgo, wu2024draganything, namekata2024sgi2v, Li_2025_ICCV} have primarily been evaluated on DAVIS~\cite{Perazzi2016}, VIPSeg~\cite{miao2021vspw}, and MagicBench~\cite{Li_2025_ICCV}. 
While existing benchmarks focus on short video sequences, FlashMotion is capable of generating videos up to 121 frames long.
This discrepancy prevents a thorough evaluation of the long-term temporal consistency and trajectory controllability of FlashMotion.
Therefore, there is an urgent need for a publicly available benchmark that targets long-sequence trajectory-controllable video generation.

Following the data pipeline introduced in MagicMotion~\cite{Li_2025_ICCV}, we build FlashBench by extending MagicBench with comprehensive trajectory annotations for all frames.
To facilitate detailed analysis, FlashBench is further organized into six groups based on the number of foreground objects, ranging from one to five, and more than five.
\section{Experiment}
We first introduce the experimental settings, including the datasets, implementation details, evaluation metrics, and comparison baselines in Sec.~\ref{sec:setup}.
Then, Sec.~\ref{sec:comparison} reports quantitative and qualitative results, conducting comprehensive comparisons with existing methods.
Finally, Sec.~\ref{sec:ablation} provides ablation studies that further analyze the contribution and effectiveness of each component of FlashMotion.

\subsection{Experiment Settings}
\label{sec:setup}
\noindent\textbf{Datasets.} 
We use MagicData~\cite{Li_2025_ICCV} as our training dataset for all the three training stages, which contains 23K high quality videos with both text and trajectory annotations, including segmentation masks and bounding boxes.
For evaluation, we conduct experiments on three different benchmarks: FlashBench, MagicBench~\cite{Li_2025_ICCV} and DAVIS~\cite{Perazzi2016}.

\noindent\textbf{Implementation details.}
In Stage1, we adopt two architectures for the trajectory adapter: ResNet~\cite{he2016deep} and ControlNet~\cite{zhang2023adding}.
The ResNet adapter is trained from scratch, while the ControlNet adapter is initialized from the main DiT weights. Both are first trained for 4.6K steps using segmentation masks as trajectory conditions, and then fine-tuned for another 5.4K steps with bounding boxes. Training is conducted on 16 A100 GPUs with a batch size of 1 per GPU and a learning rate of $2\times10^{-6}$.
In Stage 2, \fastgen is obtained by distilling Wan2.2-TI2V-5B~\cite{wan2025} into a four-step image-to-video generator. All parameters are fine-tuned for 5.5K steps on 16 A100 GPUs with a batch size of 1 per GPU. During training, the generator and fake score model are optimized with learning rates of $5\times10^{-7}$ and $1\times10^{-7}$, respectively, following a 1:5 update schedule.
In Stage 3, the trajectory adapter and discriminator are optimized with a learning rate of $2\times10^{-6}$ also under a 1:5 update ratio. The diffusion loss scale is gradually increased according to $\lambda = \tfrac{1}{4} \times 10^{-3} \times \textit{step} + 0.1$, where $\textit{step}$ denotes the current training iteration. This stage is trained for 1K steps on 4 A100 GPUs with a batch size of 1 per GPU.

\noindent\textbf{Evaluation Metrics.}
For evaluation, we follow prior works~\cite{zhou2025trackgo, wang2024motionctrl, wang2025levitor, wu2024draganything, Li_2025_ICCV} and adopt FID~\cite{heusel2017gans} and FVD~\cite{ranftl2020towards} to measure visual quality. Besides, we follow MagicMotion~\cite{Li_2025_ICCV} and employ Mask\_IoU and Box\_IoU to quantify the trajectory accuracy.

\noindent\textbf{Comparison Baselines.}
FlashMotion is evaluated against several state-of-the-art trajectory-controllable video generation methods, including MagicMotion~\cite{Li_2025_ICCV}, Tora~\cite{zhang2025tora}, DragAnything~\cite{wu2024draganything}, SGI2V~\cite{namekata2024sgi2v}, LeviTor~\cite{wang2025levitor}, and Wan2.2-TI2V-5B~\cite{wan2025} combined with the \slowadapt.
Since no existing methods support few-step trajectory-controllable video generation, we design several baselines based on existing video distillation methods~\cite{yin2024improved, luo2023latent, goodfellow2014generative} for comparison.
In these methods, we define the teacher model as the \slowadapt combined with the \slowgen, while the student model consists of the adapter paired with the \fastgen.
Since DMD~\cite{yin2024improved} and GAN~\cite{goodfellow2014generative} cause CUDA OOM errors under the ControlNet architecture, we report their results only with ResNet.

\subsection{Comparison with Other Approaches}
\label{sec:comparison}
\noindent\textbf{Quantitative comparison}

\begin{table}[t]
\caption{
Quantitative results on \textbf{FlashBench}, \textbf{MagicBench}, and \textbf{DAVIS}.
We report FID, FVD, and mask/box IoU (\%) for both ResNet and ControlNet adapters.
For each metric, the best result is highlighted in \textbf{bold}, and the second best is \underline{underlined}.
Denoising time is measured for generating 121 frames on one A100 GPU.
}
\label{tab:flashmotion_comparison}
\centering
\small
\renewcommand{\arraystretch}{1.05}
\setlength{\tabcolsep}{3pt}
\resizebox{0.96\linewidth}{!}{
\begin{tabular}{@{}lccccccccccc@{}}
\toprule
\multirow{2}{*}{\textbf{Methods}} & \multicolumn{3}{c}{\textbf{FlashBench}} & \multicolumn{3}{c}{\textbf{MagicBench}} & \multicolumn{3}{c}{\textbf{DAVIS}} & \multirow{2}{*}{\textbf{\begin{tabular}[c]{@{}c@{}}Denoising\\ Time (s)\end{tabular}}} & \multirow{2}{*}{\textbf{\begin{tabular}[c]{@{}c@{}}Params\\ (B)\end{tabular}}} \\ 
\cmidrule(lr){2-10}
 & FID($\downarrow$) & FVD($\downarrow$) & M/B IoU($\uparrow$) & FID($\downarrow$) & FVD($\downarrow$) & M/B IoU($\uparrow$) & FID($\downarrow$) & FVD($\downarrow$) & M/B IoU($\uparrow$) & & \\ 
\midrule
\multicolumn{12}{c}{\textbf{MultiSteps (50 Steps)}} \\ 
\midrule
MagicMotion~\cite{Li_2025_ICCV} & 20.03 & 138.83 & \underline{68.10/73.68} & 15.17 & \underline{107.21} & \underline{76.61/81.45} & 50.36 & 760.95 & \underline{53.94/72.84} & 1158.63 & 11.53 \\
Wan2.2 (ResNet)~\cite{wan2025} & 19.03 & 139.61 & 52.19/57.76 & 21.72 & 140.41 & 62.09/67.85 & 46.44 & \underline{703.15} & 31.22/42.74 & 333.00 & 5.02 \\
Wan2.2 (ControlNet)~\cite{wan2025} & 16.93 & 152.04 & 65.41/71.28 & 20.05 & 157.98 & 72.80/78.46 & \textbf{43.70} & 791.80 & 52.76/71.20 & 664.53 & 10.28 \\
DragAnything~\cite{wu2024draganything} & 34.93 & 267.56 & 58.54/61.72 & 31.36 & 253.40 & 66.30/70.85 & 70.70 & 1166.22 & 40.13/53.60 & 589.07 & 2.21 \\
SG-I2V~\cite{namekata2024sgi2v} & 28.52 & 252.49 & 50.20/55.72 & 32.60 & 168.82 & 68.78/74.39 & 90.93 & 1170.60 & 37.36/50.96 & 1277.15 & 1.52 \\
Tora~\cite{zhang2025tora} & 31.79 & 315.11 & 48.17/53.70 & 26.27 & 245.23 & 58.95/64.03 & 51.75 & 766.76 & 37.98/50.90 & 691.13 & 6.32 \\
LeviTor~\cite{wang2025levitor} & 64.58 & 335.47 & 36.36/39.81 & 38.32 & 194.53 & 39.96/46.36 & 97.98 & 922.68 & 25.24/31.42 & 80.08 & 2.21 \\
\midrule
\multicolumn{12}{c}{\textbf{FewSteps (4 Steps) — Adapter: ResNet}} \\ 
\midrule
DMD~\cite{yin2024improved} & 24.38 & 228.33 & 43.24/52.61 & 25.27 & 206.57 & 49.69/59.44 & 51.75 & 1058.35 & 33.08/49.78 & 11.72 & 5.02 \\
GAN~\cite{goodfellow2014generative} & 31.32 & 208.06 & 43.78/49.99 & 33.31 & 209.93 & 56.60/63.10 & 66.31 & 1143.14 & 30.49/42.80 & 11.72 & 5.02 \\
LCM~\cite{luo2023latent} & 26.79 & 462.09 & 55.31/60.80 & 28.24 & 398.06 & 64.98/70.83 & 63.07 & 1075.61 & 42.56/58.52 & 11.72 & 5.02 \\
\textbf{FlashMotion} & \underline{15.81} & \underline{108.96} & 63.96/70.01 & \underline{14.16} & 109.20 & 72.34/77.92 & 50.58 & 786.42 & 46.74/64.00 & 11.72 & 5.02 \\
\midrule
\multicolumn{12}{c}{\textbf{FewSteps (4 Steps) — Adapter: ControlNet}} \\ 
\midrule
\rowcolor{gray!10}
DMD~\cite{yin2024improved} / GAN~\cite{goodfellow2014generative} & \multicolumn{10}{c}{OOM} & -- \\
LCM~\cite{luo2023latent} & 28.34 & 340.29 & 61.29/64.83 & 25.87 & 261.87 & 70.55/74.57 & 62.25 & 1164.75 & 45.94/61.27 & 24.44 & 10.28 \\
\textbf{FlashMotion} & \textbf{14.35} & \textbf{96.08} & \textbf{69.15/75.38} & \textbf{12.49} & \textbf{99.30} & \textbf{76.92/82.17} & \underline{45.66} & \textbf{690.13} & \textbf{54.54/74.37} & 24.44 & 10.28 \\
\bottomrule
\end{tabular}
}
\end{table}

We compare FlashMotion with existing methods on FlashBench, MagicBench~\cite{Li_2025_ICCV}, and DAVIS~\cite{Perazzi2016}, evaluating both visual quality and trajectory accuracy.
In FlashBench, we use the first 121 frames of each video as the ground-truth. Since several prior methods~\cite{Li_2025_ICCV, zhang2025tora, wang2025levitor, wu2024draganything} cannot generate videos of this length, we uniformly sample $N$ frames from these 121 frames, where $N$ corresponds to the maximum video length each method supports.
In MagicBench~\cite{Li_2025_ICCV} and DAVIS~\cite{Perazzi2016}, we use the first 49 frames of each generated video for evaluation following MagicMotion~\cite{Li_2025_ICCV}.
As shown in Tab.~\ref{tab:flashmotion_comparison}, FlashMotion outperforms all existing few-step distillation methods~\cite{yin2024improved, luo2023latent, goodfellow2014generative} in both visual quality and trajectory accuracy across different adapter architectures.
When equipped with ControlNet as the adapter, FlashMotion further outperforms all prior multi-step baselines while retaining the efficiency of few-step sampling, achieving a 47× speedup over the previous SOTA~\cite{Li_2025_ICCV}.

\noindent\textbf{Qualitative comparison}

\begin{figure}[h]
    \centering
    \includegraphics[width=\linewidth]{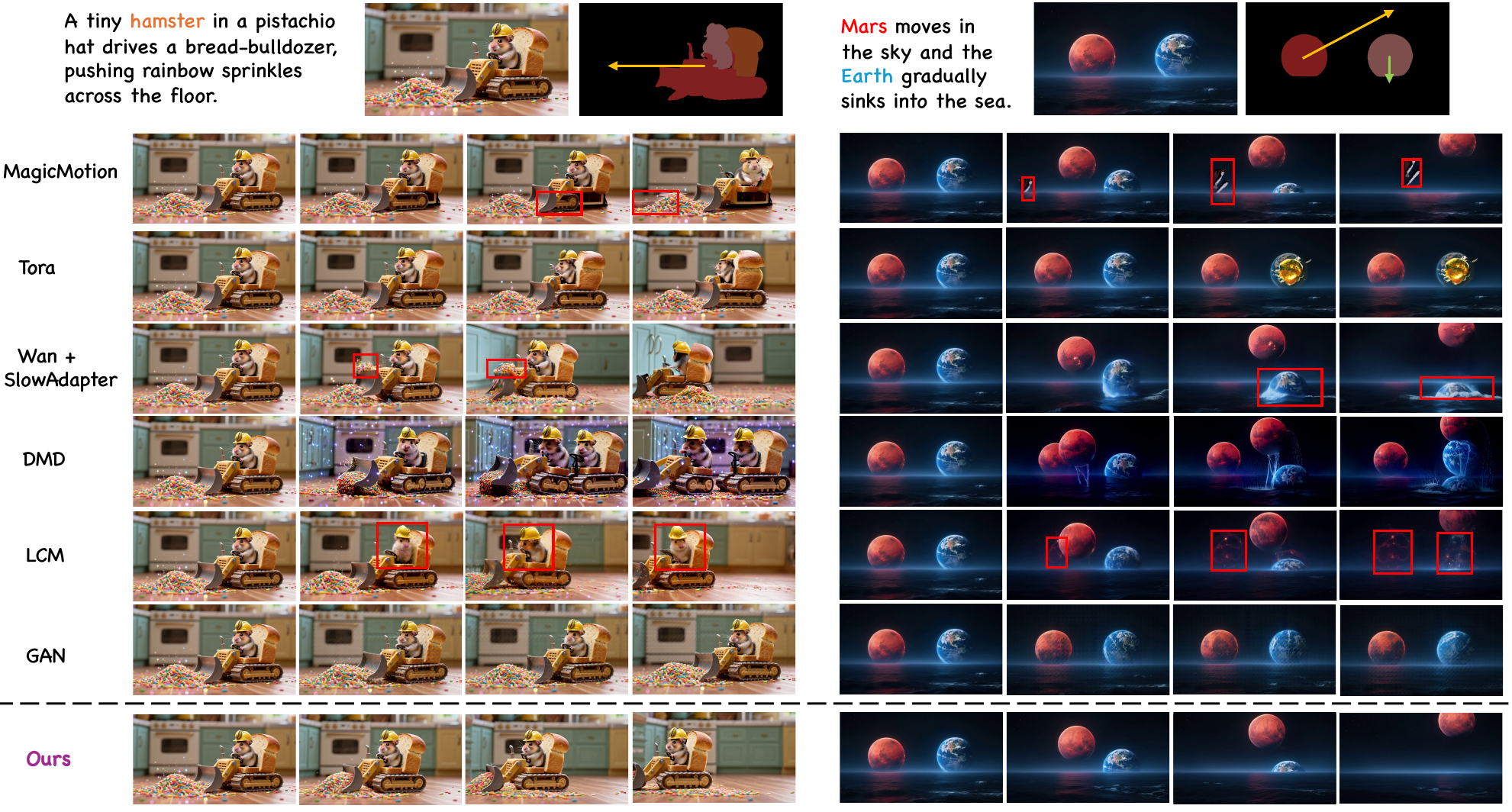}
     \caption{Qualitative Comparisons results. FlashMotion demonstrates superior qualitative performance, outperforming both previous multi-step trajectory-controllable methods and few-step distillation baselines.}
    \label{fig:comparisons}
\end{figure}

The Qualitative comparison results are presented in Fig.~\ref{fig:comparisons}, along with the corresponding input image, prompt, and trajectory. We include visualizations of all few-step baselines and four representative DiT-based multi-step baselines, MagicMotion\cite{Li_2025_ICCV}, Tora~\cite{zhang2025tora} and Wan~\cite{wan2025} + \slowgen.
As shown in Fig.~\ref{fig:comparisons}, FlashMotion outperforms all these methods on both visual quality and trajectory accuracy.

\subsection{Ablation Studies}
\label{sec:ablation}

\begin{figure}[h]
    \centering
    \includegraphics[width=0.9\linewidth]{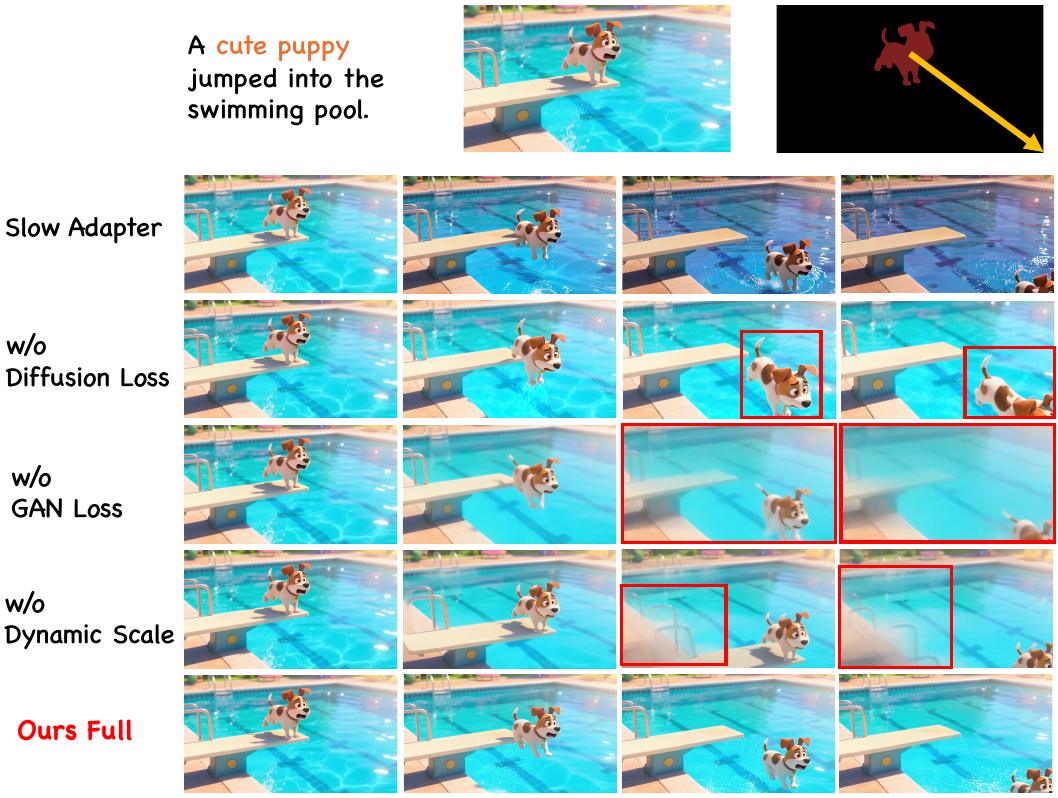}
    \caption{
    Ablation studies on the \fastadapt training stage, diffusion loss, GAN loss, and the dynamic loss scaling strategy.
    }
    \label{fig:ablation}
\end{figure}

\noindent
\begin{minipage}[c]{0.48\columnwidth}
    Due to limited space, we only present ablation results on FlashBench here in the main paper, please refer to supplementary materials for more results on MagicBench~\cite{Li_2025_ICCV} and DAVIS~\cite{Perazzi2016}.
    For fair comparison, all experiments follow the same training configurations as FlashMotion Stage3.
    
    \noindent\textbf{Fast Adapter.}
    To verify the necessity of the \fastadapt, we compute the quantitative performance of directly applying the \slowadapt to the \fastgen.
    As shown in Table.~\ref{tab:ablation_all}, removing the \fastadapt training stage leads to a notable degradation in both visual quality and trajectory accuracy.
    The result in Fig.~\ref{fig:ablation} also shows that removing this training stage can cause severe color shift in videos.
    This demonstrates that \slowadapt cannot directly control the generation process of \fastgen, highlighting the necessity of the \fastadapt training stage.

\end{minipage}
\hfill
\begin{minipage}[c]{0.5\columnwidth}
    \centering
    \captionof{table}{Ablation studies on the \fastadapt training stage, diffusion loss, GAN loss, and the dynamic loss scaling strategy.}
    \label{tab:ablation_all}
    \resizebox{\linewidth}{!}{
        \begin{tabular}{@{}lcccc@{}}
        \toprule
        \textbf{Methods} & FID$\downarrow$ & FVD$\downarrow$ & M IoU$\uparrow$ & B IoU$\uparrow$ \\ 
        \midrule
        \multicolumn{5}{c}{\textit{Adapter Type: ResNet}} \\ 
        Slow Adapter         & 22.75 & 168.46 & 49.79 & 56.62 \\
        w/o Diffusion Loss   & 18.87 & 161.07 & 52.04 & 58.04 \\
        w/o GAN Loss         & 22.74 & 206.75 & \textbf{65.82} & \textbf{70.60} \\
        w/o Dynamic Scale    & 26.32 & 210.93 & 65.54 & 69.77 \\
        \textbf{FlashMotion} & \textbf{15.81} & \textbf{108.96} & 63.96 & 70.01 \\ 
        \midrule
        \multicolumn{5}{c}{\textit{Adapter Type: ControlNet}} \\ 
        Slow Adapter         & 19.44 & 171.83 & 62.72 & 69.38 \\
        w/o Diffusion Loss   & 21.21 & 172.04 & 55.91 & 61.59 \\
        w/o GAN Loss         & 28.82 & 265.46 & \textbf{71.56} & 75.48 \\
        w/o Dynamic Scale    & 19.93 & 155.55 & 70.46 & \textbf{75.89} \\
        \textbf{FlashMotion} & \textbf{14.35} & \textbf{96.08} & 69.15 & 75.38 \\ 
        \bottomrule
        \end{tabular}
    }
\end{minipage}

\noindent\textbf{Diffusion Loss.}
We evaluate the effect of the diffusion loss by removing it during training.
As shown in Table.~\ref{tab:ablation_all} and Fig.~\ref{fig:ablation}, without the diffusion loss, the generated videos exhibit significantly lower trajectory accuracy, showing clear misalignment between the generated videos and the user-provided trajectories.
Moreover, removing the diffusion loss can also lead to decline in visual quality.

\noindent\textbf{GAN Loss.}
We perform an ablation study on the GAN loss, as shown in Table~\ref{tab:ablation_all}.
While removing the adversarial objectives slightly improves trajectory accuracy, it causes a drastic drop of nearly 90\% in visual quality, introducing severe blurring artifacts as illustrated in Fig.~\ref{fig:ablation}.

\noindent
\begin{minipage}[c]{0.48\columnwidth}
    \noindent\textbf{Dynamic Diffusion Loss Scaling.}
    We further evaluate the effectiveness of our dynamic diffusion loss scaling strategy by fixing the loss scale to 1 during training.
    As reported in Table~\ref{tab:ablation_all}, disabling the dynamic scaling mechanism leads to a noticeable decline in visual quality, again resulting in significant blurring artifacts as shown in Fig.~\ref{fig:ablation}.
    \noindent\textbf{Discriminator Architecture.}
    To validate the design of our discriminator, we conduct experiments on four different discriminator architectures.
    As shown in Table~\ref{tab:ablation_discriminator}, using only the \textit{Video Cross-Attention} layer yields the worst visual quality and trajectory accuracy.
    In contrast, incorporating the \textit{Semantic Self-Attention} module improves the model’s semantic understanding, thereby enhancing the visual quality of the generated videos, while the \textit{Trajectory Cross-Attention} module effectively strengthens trajectory control accuracy.
    Our full discriminator architecture achieves the best performance across all metrics.
\end{minipage}
\hfill
\begin{minipage}[c]{0.5\columnwidth}
    \centering
    \captionof{table}{Ablation study on the discriminator architecture on \textbf{FlashBench}. VC denotes the Video Cross-Attention layer, SS denotes the Semantic Self-Attention layer, and TC denotes the Trajectory Cross-Attention layer.}
    \label{tab:ablation_discriminator}
    \resizebox{\linewidth}{!}{
        \begin{tabular}{@{}lcccc@{}}
        \toprule
        \textbf{Methods} & FID($\downarrow$) & FVD($\downarrow$) & M IoU($\uparrow$) & B IoU($\uparrow$) \\ 
        \midrule
        \multicolumn{5}{c}{\textbf{Adapter Type: ResNet}} \\ 
        \midrule
        VC only              & 16.76 & 110.83 & 62.07 & 67.76 \\
        SS+VC                & 16.31 & 109.02 & 62.54 & 68.05 \\
        TC+VC                & 16.64 & 110.01 & 62.99 & 69.36 \\
        \textbf{FlashMotion} & \textbf{15.81} & \textbf{108.96} & \textbf{63.96} & \textbf{70.01} \\ 
        \midrule
        \multicolumn{5}{c}{\textbf{Adapter Type: ControlNet}} \\ 
        \midrule
        VC only              & 15.56 & 115.72 & 63.04 & 71.73 \\
        SS+VC                & 15.37 & 99.24  & 65.84 & 72.35 \\
        TC+VC                & 15.70 & 101.06 & 68.78 & 73.85 \\
        \textbf{FlashMotion} & \textbf{14.35} & \textbf{96.08} & \textbf{69.15} & \textbf{75.38} \\ 
        \bottomrule
        \end{tabular}
    }
\end{minipage}

\section{Conclusion}
\label{sec:conclusion}
In this work, we introduce FlashMotion, a novel framework that achieves few-step trajectory-controllable video generation through a three-stage training paradigm.
First, we train a trajectory adapter on a multi-step video generator to enable precise trajectory control.
Next, we distill the multi-step generator into a few-step version to accelerate video synthesis.
Finally, we finetune the trajectory adapter using a hybrid strategy that combines diffusion and adversarial objectives, aligning it with the few-step generator to achieve few-step trajectory-controllable video generation.
In addition, we present FlashBench, a comprehensive benchmark designed for long-sequence trajectory-controllable video generation, evaluating both visual quality and trajectory accuracy.
Extensive experiments demonstrate that FlashMotion not only surpasses existing few-step distillation approaches but also outperforms prior multi-step trajectory-controllable video generation models in both visual fidelity and trajectory consistency. \\
\textbf{Acknowledge} This work was supported by by National Natural Science Foundation of China (No. 62472098) and the Science and Technology Commission of Shanghai Municipality (No. 25511106100).

\bibliographystyle{plainnat}
\bibliography{main}

@String(CVPR= {IEEE Conf. Comput. Vis. Pattern Recog.})

@String(ICCV= {Int. Conf. Comput. Vis.})

@String(ECCV= {Eur. Conf. Comput. Vis.})

@String(NeurIPS= {Adv. Neural Inform. Process. Syst.})

@String(ICLR = {Int. Conf. Learn. Represent.})

@String(AAAI = {AAAI})

@String(CVPR  = {CVPR})

@String(ICCV  = {ICCV})

@String(ECCV  = {ECCV})

@String(NeurIPS  = {NeurIPS})

@String(ICLR  = {ICLR})

@article{ho2020denoising,
  title={Denoising diffusion probabilistic models},
  author={Ho, Jonathan and Jain, Ajay and Abbeel, Pieter},
  journal={NeurIPS},
  year={2020}
}

@article{song2020denoising,
  title={Denoising diffusion implicit models},
  author={Song, Jiaming and Meng, Chenlin and Ermon, Stefano},
  journal={arXiv preprint arXiv:2010.02502},
  year={2020}
}

@inproceedings{xing2024simda,
  title={Simda: Simple diffusion adapter for efficient video generation},
  author={Xing, Zhen and Dai, Qi and Hu, Han and Wu, Zuxuan and Jiang, Yu-Gang},
  booktitle={CVPR},
  year={2024}
}

@article{xing2024survey,
  title={A survey on video diffusion models},
  author={Xing, Zhen and Feng, Qijun and Chen, Haoran and Dai, Qi and Hu, Han and Xu, Hang and Wu, Zuxuan and Jiang, Yu-Gang},
  journal={ACM Computing Surveys},
  year={2024},
}

@article{song2020score,
  title={Score-based generative modeling through stochastic differential equations},
  author={Song, Yang and Sohl-Dickstein, Jascha and Kingma, Diederik P and Kumar, Abhishek and Ermon, Stefano and Poole, Ben},
  journal={arXiv preprint arXiv:2011.13456},
  year={2020}
}

@article{ranftl2020towards,
  title={Towards robust monocular depth estimation: Mixing datasets for zero-shot cross-dataset transfer},
  author={Ranftl, Ren{\'e} and Lasinger, Katrin and Hafner, David and Schindler, Konrad and Koltun, Vladlen},
  journal={TPAMI},
  year={2020},
  publisher={IEEE}
}

@inproceedings{zhang2023adding,
  title={Adding conditional control to text-to-image diffusion models},
  author={Zhang, Lvmin and Rao, Anyi and Agrawala, Maneesh},
  booktitle={ICCV},
  year={2023}
}

@article{vaswani2017attention,
  title={Attention is all you need},
  author={Vaswani, Ashish and Shazeer, Noam and Parmar, Niki and Uszkoreit, Jakob and Jones, Llion and Gomez, Aidan N and Kaiser, {\L}ukasz and Polosukhin, Illia},
  journal={NeurIPS},
  year={2017}
}

@article{xing2023vidiff,
  title={VIDiff: Translating Videos via Multi-Modal Instructions with Diffusion Models},
  author={Xing, Zhen and Dai, Qi and Zhang, Zihao and Zhang, Hui and Hu, Han and Wu, Zuxuan and Jiang, Yu-Gang},
  journal={arXiv preprint arXiv:2311.18837},
  year={2023}
}

@article{goodfellow2014generative,
  title={Generative adversarial nets},
  author={Goodfellow, Ian and Pouget-Abadie, Jean and Mirza, Mehdi and Xu, Bing and Warde-Farley, David and Ozair, Sherjil and Courville, Aaron and Bengio, Yoshua},
  journal={NeurIPS},
  year={2014}
}

@article{blattmann2023stable,
  title={Stable video diffusion: Scaling latent video diffusion models to large datasets},
  author={Blattmann, Andreas and Dockhorn, Tim and Kulal, Sumith and Mendelevitch, Daniel and Kilian, Maciej and Lorenz, Dominik and Levi, Yam and English, Zion and Voleti, Vikram and Letts, Adam and others},
  journal={arXiv preprint arXiv:2311.15127},
  year={2023}
}

@article{Peebles2022DiT,
  title={Scalable Diffusion Models with Transformers},
  author={William Peebles and Saining Xie},
  year={2022},
  journal={arXiv preprint arXiv:2212.09748},
}

@article{zheng2024open,
  title={Open-sora: Democratizing efficient video production for all},
  author={Zheng, Zangwei and Peng, Xiangyu and Yang, Tianji and Shen, Chenhui and Li, Shenggui and Liu, Hongxin and Zhou, Yukun and Li, Tianyi and You, Yang},
  journal={arXiv preprint arXiv:2412.20404},
  year={2024}
}

@article{yang2024cogvideox,
  title={CogVideoX: Text-to-Video Diffusion Models with An Expert Transformer},
  author={Yang, Zhuoyi and Teng, Jiayan and Zheng, Wendi and Ding, Ming and Huang, Shiyu and Xu, Jiazheng and Yang, Yuanming and Hong, Wenyi and Zhang, Xiaohan and Feng, Guanyu and others},
  journal={arXiv preprint arXiv:2408.06072},
  year={2024}
}

@article{kong2024hunyuanvideo,
  title={Hunyuanvideo: A systematic framework for large video generative models},
  author={Kong, Weijie and Tian, Qi and Zhang, Zijian and Min, Rox and Dai, Zuozhuo and Zhou, Jin and Xiong, Jiangfeng and Li, Xin and Wu, Bo and Zhang, Jianwei and others},
  journal={arXiv preprint arXiv:2412.03603},
  year={2024}
}

@inproceedings{wang2024motionctrl,
  title={Motionctrl: A unified and flexible motion controller for video generation},
  author={Wang, Zhouxia and Yuan, Ziyang and Wang, Xintao and Li, Yaowei and Chen, Tianshui and Xia, Menghan and Luo, Ping and Shan, Ying},
  booktitle={SIGGRAPH},
  year={2024}
}

@inproceedings{wu2024draganything,
  title={Draganything: Motion control for anything using entity representation},
  author={Wu, Weijia and Li, Zhuang and Gu, Yuchao and Zhao, Rui and He, Yefei and Zhang, David Junhao and Shou, Mike Zheng and Li, Yan and Gao, Tingting and Zhang, Di},
  booktitle={ECCV},
  year={2024},
  organization={Springer}
}

@article{yin2023dragnuwa,
  title={Dragnuwa: Fine-grained control in video generation by integrating text, image, and trajectory},
  author={Yin, Shengming and Wu, Chenfei and Liang, Jian and Shi, Jie and Li, Houqiang and Ming, Gong and Duan, Nan},
  journal={arXiv preprint arXiv:2308.08089},
  year={2023}
}

@inproceedings{zhang2025tora,
  title={Tora: Trajectory-oriented diffusion transformer for video generation},
  author={Zhang, Zhenghao and Liao, Junchao and Li, Menghao and Dai, Zuozhuo and Qiu, Bingxue and Zhu, Siyu and Qin, Long and Wang, Weizhi},
  booktitle={CVPR},
  year={2025}
}

@inproceedings{zhou2025trackgo,
  title={Trackgo: A flexible and efficient method for controllable video generation},
  author={Zhou, Haitao and Wang, Chuang and Nie, Rui and Liu, Jinlin and Yu, Dongdong and Yu, Qian and Wang, Changhu},
  booktitle={AAAI},
  year={2025}
}

@inproceedings{wang2025cinemaster,
  title={Cinemaster: A 3d-aware and controllable framework for cinematic text-to-video generation},
  author={Wang, Qinghe and Luo, Yawen and Shi, Xiaoyu and Jia, Xu and Lu, Huchuan and Xue, Tianfan and Wang, Xintao and Wan, Pengfei and Zhang, Di and Gai, Kun},
  booktitle={SIGGRAPH},
  year={2025}
}

@inproceedings{li2025image,
  title={Image conductor: Precision control for interactive video synthesis},
  author={Li, Yaowei and Wang, Xintao and Zhang, Zhaoyang and Wang, Zhouxia and Yuan, Ziyang and Xie, Liangbin and Shan, Ying and Zou, Yuexian},
  booktitle={AAAI},
  year={2025}
}

@inproceedings{wang2025levitor,
  title={Levitor: 3d trajectory oriented image-to-video synthesis},
  author={Wang, Hanlin and Ouyang, Hao and Wang, Qiuyu and Wang, Wen and Cheng, Ka Leong and Chen, Qifeng and Shen, Yujun and Wang, Limin},
  booktitle={CVPR},
  pages={12490--12500},
  year={2025}
}

@article{geng2024motionprompting,
  author    = {Geng, Daniel and Herrmann, Charles and Hur, Junhwa and Cole, Forrester and Zhang, Serena and Pfaff, Tobias and Lopez-Guevara, Tatiana and Doersch, Carl and Aytar, Yusuf and Rubinstein, Michael and Sun, Chen and Wang, Oliver and Owens, Andrew and Sun, Deqing},
  title     = {Motion Prompting: Controlling Video Generation with Motion Trajectories},
  journal   = {arXiv preprint arXiv:2412.02700},
  year      = {2024},
}

@inproceedings{yariv2025through,
  title={Through-The-Mask: Mask-based Motion Trajectories for Image-to-Video Generation},
  author={Yariv, Guy and Kirstain, Yuval and Zohar, Amit and Sheynin, Shelly and Taigman, Yaniv and Adi, Yossi and Benaim, Sagie and Polyak, Adam},
  booktitle={CVPR},
  year={2025}
}

@inproceedings{fu20243dtrajmaster,
  title={3DTrajMaster: Mastering 3D Trajectory for Multi-Entity Motion in Video Generation},
  author={Fu, Xiao and Liu, Xian and Wang, Xintao and Peng, Sida and Xia, Menghan and Shi, Xiaoyu and Yuan, Ziyang and Wan, Pengfei and Zhang, Di and Lin, Dahua},
  booktitle={ICLR},
  year={2025}
}

@article{gu2025das,
    title={Diffusion as Shader: 3D-aware Video Diffusion for Versatile Video Generation Control}, 
    author={Zekai Gu and Rui Yan and Jiahao Lu and Peng Li and Zhiyang Dou and Chenyang Si and Zhen Dong and Qifeng Liu and Cheng Lin and Ziwei Liu and Wenping Wang and Yuan Liu},
    year={2025},
    journal={arXiv preprint arXiv:2501.03847}
}

@inproceedings{objctrl2.5d,
  title={{ObjCtrl-2.5D}: Training-free Object Control with Camera Poses},
  author={Wang, Zhouxia and Lan, Yushi and Zhou, Shangchen and Loy, Chen Change},
  booktitle={arXiv preprint arXiv:2412.07721},
  year={2024}
}

@article{qiu2024freetraj,
  title={Freetraj: Tuning-free trajectory control in video diffusion models},
  author={Qiu, Haonan and Chen, Zhaoxi and Wang, Zhouxia and He, Yingqing and Xia, Menghan and Liu, Ziwei},
  journal={arXiv preprint arXiv:2406.16863},
  year={2024}
}

@inproceedings{yang2024direct,
  title={Direct-a-video: Customized video generation with user-directed camera movement and object motion},
  author={Yang, Shiyuan and Hou, Liang and Huang, Haibin and Ma, Chongyang and Wan, Pengfei and Zhang, Di and Chen, Xiaodong and Liao, Jing},
  booktitle={SIGGRAPH},
  year={2024}
}

@inproceedings{namekata2024sgi2v,
  title = {SG-I2V: Self-Guided Trajectory Control in Image-to-Video Generation},
  author = {Namekata, Koichi and Bahmani, Sherwin and Wu, Ziyi and Kant, Yash and Gilitschenski, Igor and Lindell, David B.},
  booktitle = {ICLR},
  year = {2025},
  url = {https://openreview.net/forum?id=uQjySppU9x},
}

@article{jain2023peekaboo,
  title={PEEKABOO: Interactive Video Generation via Masked-Diffusion},
  author={Jain, Yash and Nasery, Anshul and Vineet, Vibhav and Behl, Harkirat},
  journal={arXiv preprint arXiv:2312.07509},
  year={2023}
}

@article{shi2024motion,
    title={Motion-i2v: Consistent and controllable image-to-video generation with explicit motion modeling},
    author={Shi, Xiaoyu and Huang, Zhaoyang and Wang, Fu-Yun and Bian, Weikang and Li, Dasong and Zhang, Yi and Zhang, Manyuan and Cheung, Ka Chun and See, Simon and Qin, Hongwei and others},
    journal={SIGGRAPH},
    year={2024}
}

@inproceedings{miao2021vspw,
  title={Vspw: A large-scale dataset for video scene parsing in the wild},
  author={Miao, Jiaxu and Wei, Yunchao and Wu, Yu and Liang, Chen and Li, Guangrui and Yang, Yi},
  booktitle={CVPR},
  year={2021}
}

@inproceedings{Perazzi2016,
  author = {F. Perazzi and J. Pont-Tuset and B. McWilliams and L. {Van Gool} and M. Gross and A. Sorkine-Hornung},
  title = {A Benchmark Dataset and Evaluation Methodology for Video Object Segmentation},
  booktitle = {CVPR},
  year = {2016}
}

@article{kingma2013auto,
      title={Auto-Encoding Variational Bayes}, 
      author={Diederik P Kingma and Max Welling},
      journal={arXiv preprint arXiv:1312.6114},
      year={2013}
}

@article{heusel2017gans,
  title={Gans trained by a two time-scale update rule converge to a local nash equilibrium},
  author={Heusel, Martin and Ramsauer, Hubert and Unterthiner, Thomas and Nessler, Bernhard and Hochreiter, Sepp},
  journal={NeurIPS},
  year={2017}
}

@article{wan2025,
      title={Wan: Open and Advanced Large-Scale Video Generative Models}, 
      author={Team Wan and Ang Wang and Baole Ai and Bin Wen and Chaojie Mao and Chen-Wei Xie and Di Chen and Feiwu Yu and Haiming Zhao and Jianxiao Yang and Jianyuan Zeng and Jiayu Wang and Jingfeng Zhang and Jingren Zhou and Jinkai Wang and Jixuan Chen and Kai Zhu and Kang Zhao and Keyu Yan and Lianghua Huang and Mengyang Feng and Ningyi Zhang and Pandeng Li and Pingyu Wu and Ruihang Chu and Ruili Feng and Shiwei Zhang and Siyang Sun and Tao Fang and Tianxing Wang and Tianyi Gui and Tingyu Weng and Tong Shen and Wei Lin and Wei Wang and Wei Wang and Wenmeng Zhou and Wente Wang and Wenting Shen and Wenyuan Yu and Xianzhong Shi and Xiaoming Huang and Xin Xu and Yan Kou and Yangyu Lv and Yifei Li and Yijing Liu and Yiming Wang and Yingya Zhang and Yitong Huang and Yong Li and You Wu and Yu Liu and Yulin Pan and Yun Zheng and Yuntao Hong and Yupeng Shi and Yutong Feng and Zeyinzi Jiang and Zhen Han and Zhi-Fan Wu and Ziyu Liu},
      journal = {arXiv preprint arXiv:2503.20314},
      year={2025}
}

@article{xing2024aid,
  title={Aid: Adapting image2video diffusion models for instruction-guided video prediction},
  author={Xing, Zhen and Dai, Qi and Weng, Zejia and Wu, Zuxuan and Jiang, Yu-Gang},
  journal={arXiv preprint arXiv:2406.06465},
  year={2024}
}

@article{huang2025selfforcing,
  title={Self Forcing: Bridging the Train-Test Gap in Autoregressive Video Diffusion},
  author={Huang, Xun and Li, Zhengqi and He, Guande and Zhou, Mingyuan and Shechtman, Eli},
  journal={arXiv preprint arXiv:2506.08009},
  year={2025}
}

@inproceedings{yin2024improved,
    title={Improved Distribution Matching Distillation for Fast Image Synthesis},
    author={Yin, Tianwei and Gharbi, Micha{\"e}l and Park, Taesung and Zhang, Richard and Shechtman, Eli and Durand, Fredo and Freeman, William T},
    booktitle={NeurIPS},
    year={2024}
}

@inproceedings{yin2024onestep,
    title={One-step Diffusion with Distribution Matching Distillation},
    author={Yin, Tianwei and Gharbi, Micha{\"e}l and Zhang, Richard and Shechtman, Eli and Durand, Fr{\'e}do and Freeman, William T and Park, Taesung},
    booktitle={CVPR},
    year={2024}
}

@InProceedings{Li_2025_ICCV,
    author    = {Li, Quanhao and Xing, Zhen and Wang, Rui and Zhang, Hui and Dai, Qi and Wu, Zuxuan},
    title     = {MagicMotion: Controllable Video Generation with Dense-to-Sparse Trajectory Guidance},
    booktitle = {ICCV},
    year      = {2025},
}

@InProceedings{He_2016_CVPR,
author = {He, Kaiming and Zhang, Xiangyu and Ren, Shaoqing and Sun, Jian},
title = {Deep Residual Learning for Image Recognition},
booktitle = {CVPR},
year = {2016}
}

@article{lin2025diffusion,
  title={Diffusion adversarial post-training for one-step video generation},
  author={Lin, Shanchuan and Xia, Xin and Ren, Yuxi and Yang, Ceyuan and Xiao, Xuefeng and Jiang, Lu},
  journal={arXiv preprint arXiv:2501.08316},
  year={2025}
}

@article{luo2023latent,
  title={Latent consistency models: Synthesizing high-resolution images with few-step inference},
  author={Luo, Simian and Tan, Yiqin and Huang, Longbo and Li, Jian and Zhao, Hang},
  journal={arXiv preprint arXiv:2310.04378},
  year={2023}
}

@article{zhang2025tora2,
  title={Tora2: Motion and Appearance Customized Diffusion Transformer for Multi-Entity Video Generation},
  author={Zhang, Zhenghao and Liao, Junchao and Meng, Xiangyu and Qin, Long and Wang, Weizhi},
  journal={arXiv preprint arXiv:2507.05963},
  year={2025}
}

@article{wang2023videolcm,
  title={Videolcm: Video latent consistency model},
  author={Wang, Xiang and Zhang, Shiwei and Zhang, Han and Liu, Yu and Zhang, Yingya and Gao, Changxin and Sang, Nong},
  journal={arXiv preprint arXiv:2312.09109},
  year={2023}
}

@article{song2023consistency,
  title={Consistency Models},
  author={Song, Yang and Dhariwal, Prafulla and Chen, Mark and Sutskever, Ilya},
  journal={arXiv preprint arXiv:2303.01469},
  year={2023},
}

@inproceedings{yin2025causvid,
    title={From Slow Bidirectional to Fast Autoregressive Video Diffusion Models},
    author={Yin, Tianwei and Zhang, Qiang and Zhang, Richard and Freeman, William T and Durand, Fredo and Shechtman, Eli and Huang, Xun},
    booktitle={CVPR},
    year={2025}
}

@article{lin2025autoregressive,
  title={Autoregressive Adversarial Post-Training for Real-Time Interactive Video Generation},
  author={Lin, Shanchuan and Yang, Ceyuan and He, Hao and Jiang, Jianwen and Ren, Yuxi and Xia, Xin and Zhao, Yang and Xiao, Xuefeng and Jiang, Lu},
  journal={arXiv preprint arXiv:2506.09350},
  year={2025}
}

@inproceedings{ma2024trailblazer,
  title={Trailblazer: Trajectory control for diffusion-based video generation},
  author={Ma, Wan-Duo Kurt and Lewis, John P and Kleijn, W Bastiaan},
  booktitle={SIGGRAPH Asia 2024 Conference Papers},
  year={2024}
}

@article{li2024t2v,
  title={T2v-turbo: Breaking the quality bottleneck of video consistency model with mixed reward feedback},
  author={Li, Jiachen and Feng, Weixi and Fu, Tsu-Jui and Wang, Xinyi and Basu, Sugato and Chen, Wenhu and Wang, William Yang},
  journal={Advances in neural information processing systems},
  year={2024}
}

@article{lv2025dcm,
  title={DCM: Dual-Expert Consistency Model for Efficient and High-Quality Video Generation},
  author={Lv, Zhengyao and Si, Chenyang and Pan, Tianlin and Chen, Zhaoxi and Wong, Kwan-Yee K and Qiao, Yu and Liu, Ziwei},
  journal={arXiv preprint arXiv:2506.03123},
  year={2025}
}

@inproceedings{sauer2024adversarial,
  title={Adversarial diffusion distillation},
  author={Sauer, Axel and Lorenz, Dominik and Blattmann, Andreas and Rombach, Robin},
  booktitle={European Conference on Computer Vision},
  year={2024},
}

@inproceedings{sauer2024fast,
  title={Fast high-resolution image synthesis with latent adversarial diffusion distillation},
  author={Sauer, Axel and Boesel, Frederic and Dockhorn, Tim and Blattmann, Andreas and Esser, Patrick and Rombach, Robin},
  booktitle={SIGGRAPH Asia 2024 Conference Papers},
  year={2024}
}

@article{shao2025magicdistillation,
  title={MagicDistillation: Weak-to-Strong Video Distillation for Large-Scale Few-Step Synthesis},
  author={Shao, Shitong and Yi, Hongwei and Guo, Hanzhong and Ye, Tian and Zhou, Daquan and Lingelbach, Michael and Xu, Zhiqiang and Xie, Zeke},
  journal={arXiv preprint arXiv:2503.13319},
  year={2025}
}

@article{cheng2025pose,
  title={POSE: Phased One-Step Adversarial Equilibrium for Video Diffusion Models},
  author={Cheng, Jiaxiang and Ma, Bing and Ren, Xuhua and Jin, Hongyi and Yu, Kai and Zhang, Peng and Li, Wenyue and Zhou, Yuan and Zheng, Tianxiang and Lu, Qinglin},
  journal={arXiv preprint arXiv:2508.21019},
  year={2025}
}

@inproceedings{he2016deep,
  title={Deep residual learning for image recognition},
  author={He, Kaiming and Zhang, Xiangyu and Ren, Shaoqing and Sun, Jian},
  booktitle={Proceedings of the IEEE conference on computer vision and pattern recognition},
  year={2016}
}

@article{sun2025swiftvideo,
  title={Swiftvideo: A unified framework for few-step video generation through trajectory-distribution alignment},
  author={Sun, Yanxiao and Wu, Jiafu and Cao, Yun and Xu, Chengming and Wang, Yabiao and Cao, Weijian and Luo, Donghao and Wang, Chengjie and Fu, Yanwei},
  journal={arXiv preprint arXiv:2508.06082},
  year={2025}
}

@article{HaCohen2024LTXVideo,
  title={LTX-Video: Realtime Video Latent Diffusion},
  author={HaCohen, Yoav and Chiprut, Nisan and Brazowski, Benny and Shalem, Daniel and Moshe, Dudu and Richardson, Eitan and Levin, Eran and Shiran, Guy and Zabari, Nir and Gordon, Ori and Panet, Poriya and Weissbuch, Sapir and Kulikov, Victor and Bitterman, Yaki and Melumian, Zeev and Bibi, Ofir},
  journal={arXiv preprint arXiv:2501.00103},
  year={2024}
}

@inproceedings{NIPS2000_44968aec,
 author = {Dugas, Charles and Bengio, Yoshua and B\'{e}lisle, Fran\c{c}ois and Nadeau, Claude and Garcia, Ren\'{e}},
 booktitle = {Advances in Neural Information Processing Systems},
 editor = {T. Leen and T. Dietterich and V. Tresp},
 title = {Incorporating Second-Order Functional Knowledge for Better Option Pricing},
 year = {2000}
}

@inproceedings{wei2024dreamvideo,
  title={Dreamvideo: Composing your dream videos with customized subject and motion},
  author={Wei, Yujie and Zhang, Shiwei and Qing, Zhiwu and Yuan, Hangjie and Liu, Zhiheng and Liu, Yu and Zhang, Yingya and Zhou, Jingren and Shan, Hongming},
  booktitle={Proceedings of the IEEE/CVF Conference on Computer Vision and Pattern Recognition},
  year={2024}
}

@inproceedings{wei2025dreamrelation,
  title={Dreamrelation: Relation-centric video customization},
  author={Wei, Yujie and Zhang, Shiwei and Yuan, Hangjie and Gong, Biao and Tang, Longxiang and Wang, Xiang and Qiu, Haonan and Li, Hengjia and Tan, Shuai and Zhang, Yingya and others},
  booktitle={Proceedings of the IEEE/CVF International Conference on Computer Vision},
  year={2025}
}

@article{wei2024dreamvideo2,
  title={Dreamvideo-2: Zero-shot subject-driven video customization with precise motion control},
  author={Wei, Yujie and Zhang, Shiwei and Yuan, Hangjie and Wang, Xiang and Qiu, Haonan and Zhao, Rui and Feng, Yutong and Liu, Feng and Huang, Zhizhong and Ye, Jiaxin and others},
  journal={arXiv preprint arXiv:2410.13830},
  year={2024}
}

\clearpage
\section*{Appendix}
\section{Additional Ablation results}
\subsection{Quantitative Results}
Here, we provide the complete quantitative results across all three benchmarks, including FlashBench, MagicBench~\cite{Li_2025_ICCV}, and DAVIS~\cite{Perazzi2016} in Table.~\ref{tab:ablation_all_supp} and Table.~\ref{tab:ablation_discriminator_supp}.
All ablation studies are trained for 1K steps on 4 Nvidia A100 GPUs, with other training configurations kept consistent with FlashMotion Stage 3.

\begin{table*}[h]
\caption{Comprehensive ablation study of FlashMotion. We analyze both adapter variants (ResNet and ControlNet) by progressively removing key components — including the \fastadapt training stage, diffusion loss, GAN loss, and the dynamic diffusion loss scaling strategy. The results show that each component plays a crucial role in preserving high video quality and precise motion alignment.}
\label{tab:ablation_all_supp}
\centering
\small
\renewcommand{\arraystretch}{1.1}
\setlength{\tabcolsep}{4pt}
\resizebox{\linewidth}{!}{
\begin{tabular}{@{}lccccccccc@{}}
\toprule
\multirow{2}{*}{\textbf{Methods}} & \multicolumn{3}{c}{\textbf{FlashBench}}                        & \multicolumn{3}{c}{\textbf{MagicBench}}                        & \multicolumn{3}{c}{\textbf{DAVIS}}                             \\ \cmidrule(l){2-10} 
                                  & FID($\downarrow$) & FVD($\downarrow$) & M/B IoU\%($\uparrow$)  & FID($\downarrow$) & FVD($\downarrow$) & M/B IoU\%($\uparrow$)  & FID($\downarrow$) & FVD($\downarrow$) & M/B IoU\%($\uparrow$)  \\ \midrule
\multicolumn{10}{c}{\textbf{Adapter Type: ResNet}}                                                                                                                                                                                   \\ \midrule
Slow Adapter                      & 22.75             & 168.46            & 49.79 / 56.62          & 21.59             & 162.93            & 60.24 / 67.23          & 52.01             & 992.26            & 36.33 / 51.37          \\
w/o Diffusion Loss                & 18.87             & 161.07            & 52.04 / 58.04          & 21.95             & 162.31            & 63.14 / 69.02          & 55.28             & 983.91            & 37.22 / 52.47          \\
w/o GAN Loss                      & 22.74             & 206.75            & \textbf{65.82 / 70.60} & 30.51             & 167.91            & \textbf{73.86 / 78.48} & 66.46             & 1015.81           & \textbf{47.13} / 62.58          \\
w/o Dynamic Scale                 & 26.32             & 210.93            & 65.54 / 69.77          & 21.90             & 167.00            & 73.60 / 78.15          & 73.12             & 998.85            & 47.01 / 60.12          \\
\textbf{FlashMotion}              & \textbf{15.81}    & \textbf{108.96}   & 63.96 / 70.01          & \textbf{14.16}    & \textbf{109.20}   & 72.34 / 77.92          & \textbf{50.58}    & \textbf{786.42}   & 46.74 / \textbf{64.00} \\ \midrule
\multicolumn{10}{c}{\textbf{Adapter Type: ControlNet}}                                                                                                                                                                               \\ \midrule
Slow Adapter                      & 19.44             & 171.83            & 62.72 / 69.38          & 21.19             & 161.80            & 70.20 / 76.54          & 46.42             & 875.37            & 50.52 / 70.83          \\
w/o Diffusion Loss                & 21.21             & 172.04            & 55.91 / 61.59          & 22.36             & 176.01            & 66.25 / 71.82          & 49.27             & 882.81            & 42.46 / 59.01          \\
w/o GAN Loss                      & 28.82             & 265.46            & \textbf{71.56} / 75.48 & 26.33             & 192.85            & \textbf{78.26} / 82.15 & 75.42             & 1131.65           & \textbf{55.87} / 68.59          \\
w/o Dynamic Scale                 & 19.93             & 155.55            & 70.46 / \textbf{75.89}          & 16.83             & 131.59            & 77.49 / \textbf{82.29}          & 61.47             & 958.22            & 55.51 / 70.13          \\
\textbf{FlashMotion}              & \textbf{14.35}    & \textbf{96.08}    & 69.15 / 75.38          & \textbf{12.49}    & \textbf{99.30}    & 76.92 / 82.17          & \textbf{45.66}    & \textbf{690.13}   & 54.54 / \textbf{74.37} \\ \bottomrule
\end{tabular}
}
\end{table*}
\begin{table*}[h]
\caption{
Ablation study on the discriminator architecture.
VC denotes the \textit{Video Cross-Attention} layer, SS denotes the \textit{Semantic Self-Attention} layer, and TC denotes the \textit{Trajectory Cross-Attention} layer.
Results show that our discriminator design achieves the best overall performance across all benchmarks and metrics.
}
\label{tab:ablation_discriminator_supp}
\centering
\small
\renewcommand{\arraystretch}{1.1}
\setlength{\tabcolsep}{4pt}
\resizebox{\linewidth}{!}{
\begin{tabular}{@{}lccccccccc@{}}
\toprule
\multirow{2}{*}{\textbf{Methods}} & \multicolumn{3}{c}{\textbf{FlashBench}}                        & \multicolumn{3}{c}{\textbf{MagicBench}}                        & \multicolumn{3}{c}{\textbf{DAVIS}}                             \\ \cmidrule(l){2-10} 
                                  & FID($\downarrow$) & FVD($\downarrow$) & M/B IoU\%($\uparrow$)  & FID($\downarrow$) & FVD($\downarrow$) & M/B IoU\%($\uparrow$)  & FID($\downarrow$) & FVD($\downarrow$) & M/B IoU\%($\uparrow$)  \\ \midrule
\multicolumn{10}{c}{\textbf{Adapter Type: ResNet}}                                                                                                                                                                                   \\ \midrule
VC only                           & 16.76             & 110.83            & 62.07 / 67.76          & 14.73             & 114.61            & 71.00 / 75.86          & 53.22             & 800.50            & 43.97 / 60.16          \\
SS+VC                             & 16.31             & 109.02            & 62.54 / 68.05          & 14.44             & 113.88            & 71.16 / 76.28          & 52.34             & 830.14            & 44.61 / 62.50          \\
TC+VC                             & 16.64             & 110.01            & 62.99 / 69.36          & 14.87             & 114.11            & 71.70 / 77.31          & 53.16             & 830.57            & 45.11 / 62.56          \\
\textbf{FlashMotion}              & \textbf{15.81}    & \textbf{108.96}   & \textbf{63.96 / 70.01} & \textbf{14.16}    & \textbf{109.20}   & \textbf{72.34 / 77.92} & \textbf{50.58}    & \textbf{786.42}   & \textbf{46.74 / 64.00} \\ \midrule
\multicolumn{10}{c}{\textbf{Adapter Type: ControlNet}}                                                                                                                                                                               \\ \midrule
VC only                           & 15.56             & 115.72            & 63.04 / 71.73          & 13.71             & 120.22            & 75.78 / 81.33          & 49.39             & 798.79            & 51.48 / 69.00          \\
SS+VC                             & 15.37             & 99.24             & 65.84 / 72.35          & 13.42             & 101.58            & 75.35 / 81.06          & 46.24             & 711.82            & 53.33 / 71.99          \\
TC+VC                             & 15.70             & 101.06            & 68.78 / 73.85          & 13.96             & 105.49            & 76.48 / 82.15          & 48.50             & 758.96            & 53.91 / 72.90          \\
\textbf{FlashMotion}              & \textbf{14.35}    & \textbf{96.08}        & \textbf{69.15 / 75.38} & \textbf{12.49}    & \textbf{99.30}    & \textbf{76.92 / 82.17} & \textbf{45.66}    & \textbf{690.13}   & \textbf{54.54 / 74.37} \\ \bottomrule
\end{tabular}
}
\end{table*}

\paragraph{Fast Adapter}
To assess the importance of the \fastadapt training stage, we evaluate the performance of directly applying \slowadapt to \fastgen across all three benchmarks.
As shown in Table~\ref{tab:ablation_all_supp}, removing the \fastadapt stage results in a consistent decline in both video quality and trajectory accuracy across all benchmarks, underscoring the necessity of the additional \fastadapt training stage.
\paragraph{Diffusion Loss}
To evaluate the role of the diffusion loss, we remove it during training and measure performance across all benchmarks.
As presented in Table~\ref{tab:ablation_all_supp}, removing the diffusion loss leads to a noticeable drop in trajectory alignment for both adapter architectures.
This shows that the diffusion loss is essential for maintaining trajectory consistency between generated motions and user-specified trajectories.
Moreover, its removal also causes a degradation in both image and video quality.
\paragraph{GAN Loss}
We conduct an ablation study on the GAN loss, as summarized in Table~\ref{tab:ablation_all_supp}.
While removing the adversarial objectives slightly improves trajectory accuracy, it causes an approximately 90\% reduction in both image and video quality, introducing severe blurring artifacts.
\paragraph{Dynamic Diffusion Loss Scaling}
We further validate the effectiveness of the proposed dynamic diffusion loss scaling strategy by fixing the loss scale to 1 during training.
As shown in Table~\ref{tab:ablation_all_supp}, disabling dynamic scaling leads to a clear decline in both image and video quality across all three benchmarks, again resulting in noticeable blurring artifacts.
\paragraph{Discriminator Architecture}
Finally, we assess the impact of different discriminator architectures, as shown in Table~\ref{tab:ablation_discriminator_supp}.
Using only the \textit{Video Cross-Attention} layer yields the lowest performance in both visual quality and trajectory accuracy.
In contrast, incorporating the \textit{Semantic Self-Attention} module enhances the model’s semantic understanding, improving visual realism, while the \textit{Trajectory Cross-Attention} module strengthens trajectory control accuracy.
Overall, our full discriminator architecture achieves the best results across all evaluation metrics and benchmarks.

\subsection{More Qualitative Results}
\begin{figure*}[ht]
    \centering
    \includegraphics[width=\linewidth]{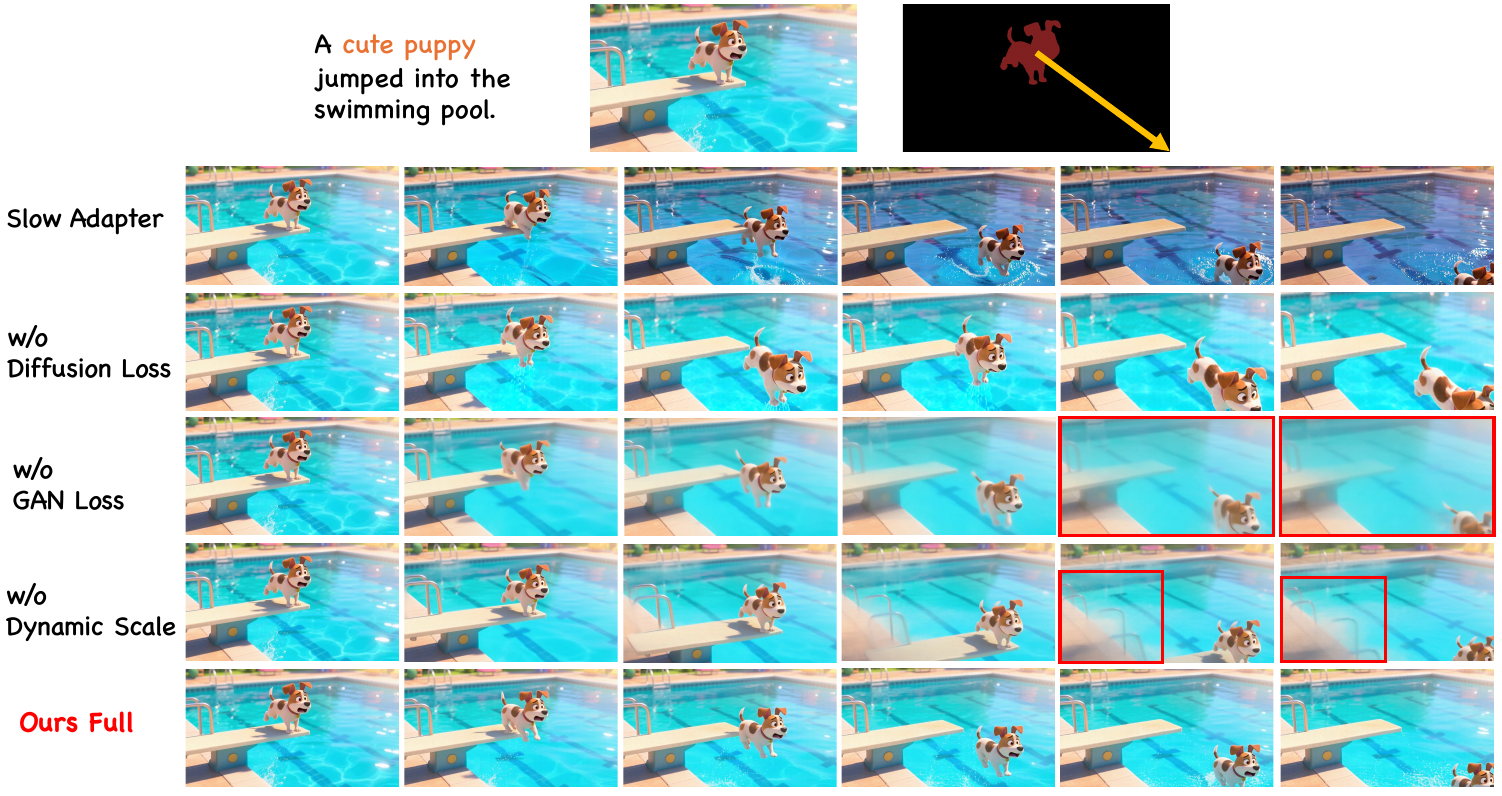}
     \caption{Additional ablation study results. Only our full method can generate videos with both high visual quality and trajectory accuracy.}
    \label{fig:Ablations_supp1}
\end{figure*}
\begin{figure*}[ht]
    \centering
    \includegraphics[width=\linewidth]{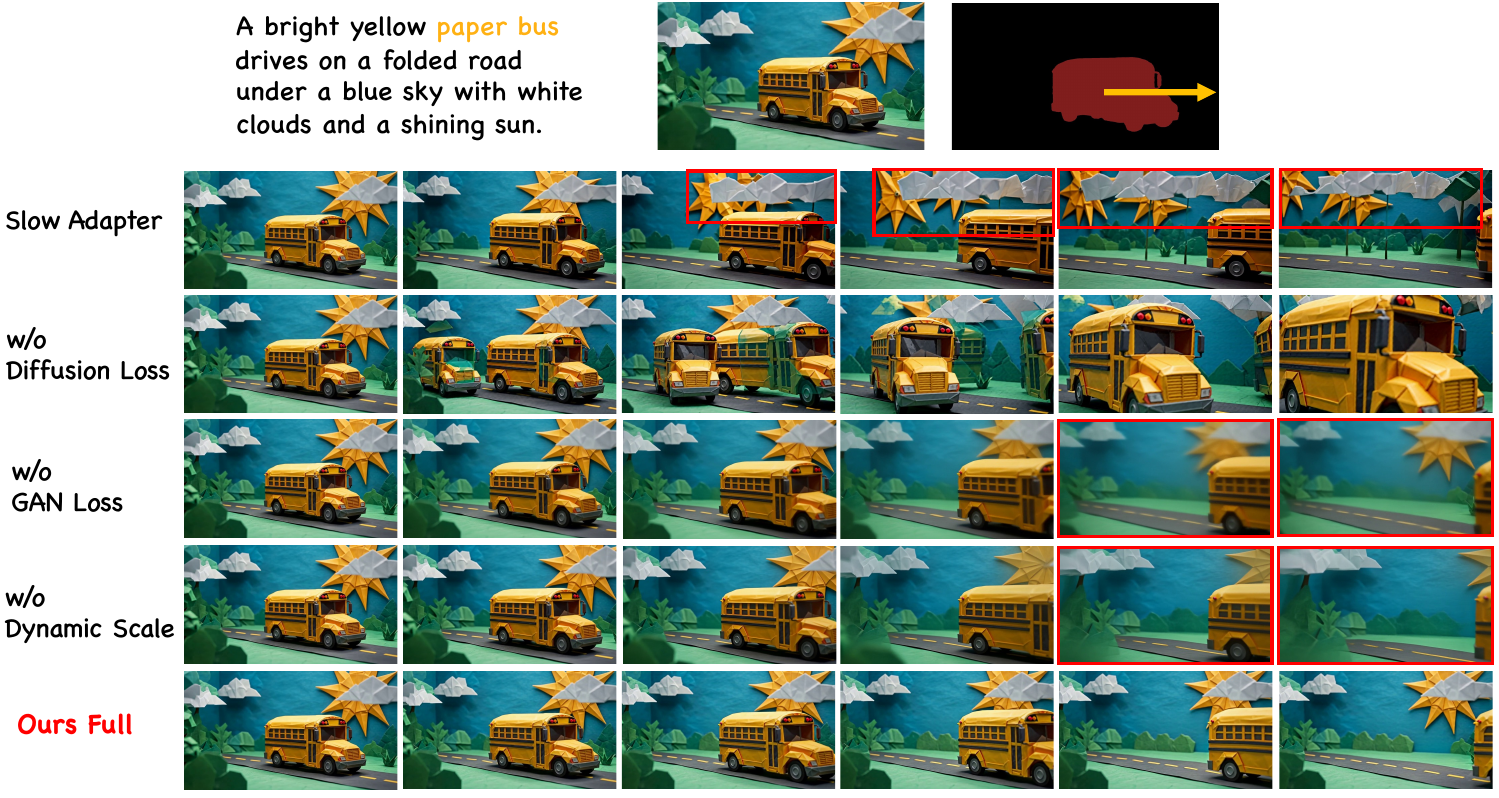}
     \caption{Additional ablation study results. Only our full method can generate videos with both high visual quality and trajectory accuracy.}
    \label{fig:Ablations_supp2}
\end{figure*}
\begin{figure*}[ht]
    \centering
    \includegraphics[width=\linewidth]{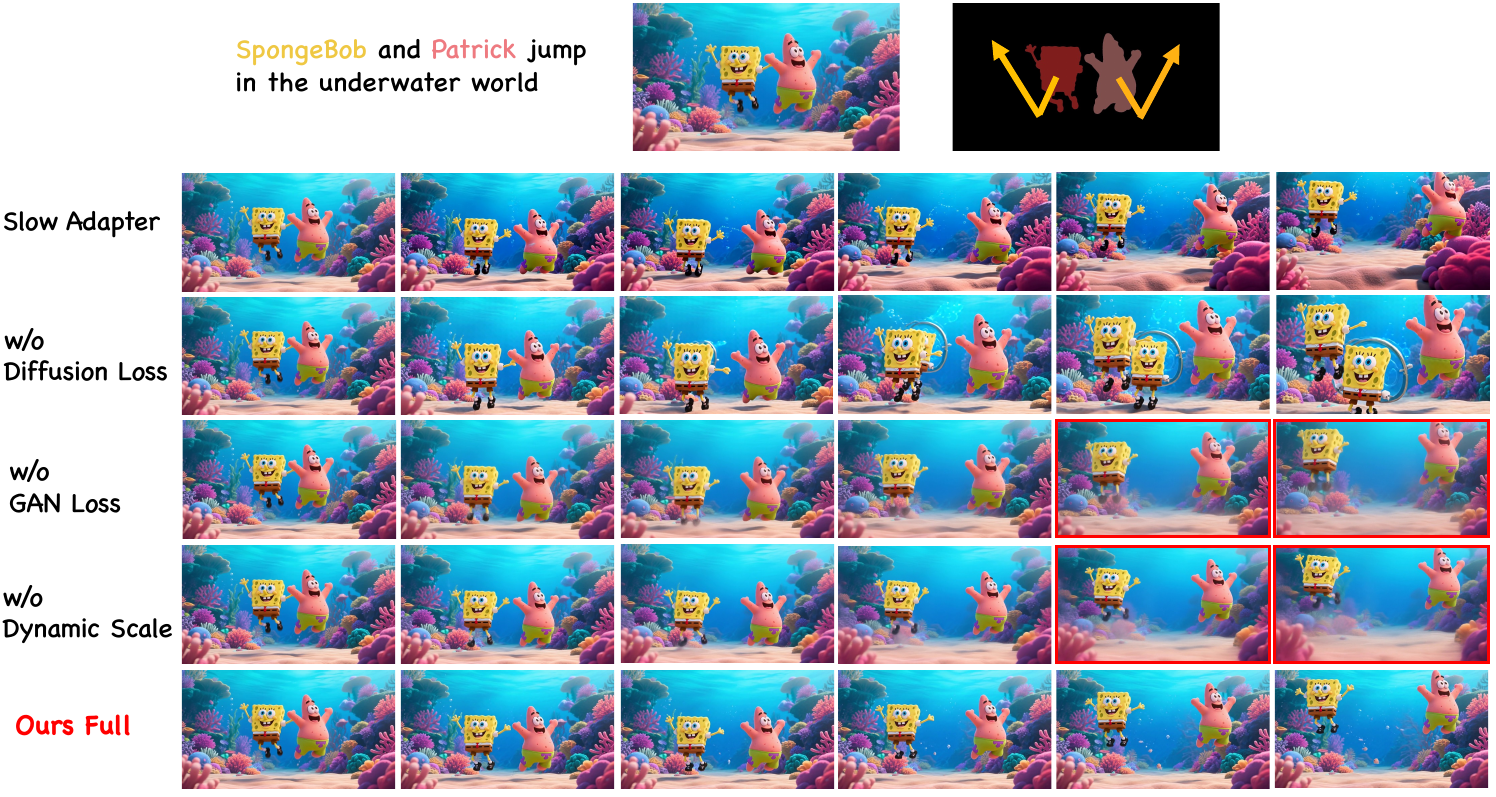}
     \caption{Additional ablation study results. Only our full method can generate videos with both high visual quality and trajectory accuracy.}
    \label{fig:Ablations_supp3}
\end{figure*}

Detailed qualitative ablation results are presented in Fig.\ref{fig:Ablations_supp1}, Fig.\ref{fig:Ablations_supp2}, and Fig.\ref{fig:Ablations_supp3}.
As shown, directly applying \slowadapt to \fastgen produces pronounced artifacts—such as the color drift in Fig.\ref{fig:Ablations_supp1} and Fig.\ref{fig:Ablations_supp3}, and the distorted object shapes in Fig.\ref{fig:Ablations_supp2}.
In addition, removing the diffusion loss during training markedly degrades trajectory fidelity: objects (e.g., the dog or the bus) drift away from the intended paths, and in the extreme case shown in Fig.~\ref{fig:Ablations_supp3}, a single Spongebob is mistakenly duplicated into two.
Finally, eliminating either the GAN loss or the dynamic scale strategy introduces severe blurring artifacts.
\section{Additional Comparison results}
\subsection{Backbone Comparisons}
As shown in Table~\ref{tab:backbone}, we present a comprehensive comparison of the backbone architectures used across different methods. The table summarizes the supported video length and spatial resolution, as well as the corresponding denoising latency and total parameter count.
Notably, FlashMotion achieves the fastest denoising speed for both the ControlNet- and ResNet-based adapters, while also supporting the highest resolution and the longest generation length. Depending on their needs, users can flexibly choose between the ResNet or ControlNet variants of FlashMotion to balance generation speed, video quality, and trajectory accuracy.

\begin{table*}[h]
\caption{Comparison of model configurations and backbone architectures, including supported video length, spatial resolution, denoising latency, and total parameters. FlashMotion achieves the fastest denoising speed while supporting the highest resolution and longest generation length.}
\centering
\label{tab:backbone}
\resizebox{\linewidth}{!}{
\begin{tabular}{@{}cccccc@{}}
\toprule
\textbf{Method} & Video Length & Video Resolution & Denoising Latency(s) & Total Params(B) & Base Model \\ 
\midrule
LeviTor~\cite{wang2025levitor}         & 16  & 288$\times$512  & 80.08  & 2.21  & SVD~\cite{blattmann2023stable} \\
DragAnything~\cite{wu2024draganything} & 14  & 320$\times$576  & 589.07 & 2.21  & SVD~\cite{blattmann2023stable} \\
SG-I2V~\cite{namekata2024sgi2v}        & 14  & 576$\times$1024 & 1277.15 & 1.52 & SVD~\cite{blattmann2023stable} \\
Tora~\cite{zhang2025tora}              & 49  & 480$\times$720  & 691.13 & 6.32 & CogVideoX~\cite{yang2024cogvideox} \\
MagicMotion~\cite{Li_2025_ICCV}   & 49  & 480$\times$720  & 1158.63 & 11.53 & CogVideoX~\cite{yang2024cogvideox} \\
Wan+ResNet~\cite{wan2025}              & 121 & 704$\times$1280 & 333.00 & 5.02 & Wan2.2~\cite{wan2025} \\
Wan+ControlNet~\cite{wan2025}          & 121 & 704$\times$1280 & 664.53 & 10.28 & Wan2.2~\cite{wan2025} \\
\textbf{FlashMotion (ResNet)}          & 121 & 704$\times$1280 & 11.72 & 5.02 & Wan2.2~\cite{wan2025} \\
\textbf{FlashMotion (ControlNet)}      & 121 & 704$\times$1280 & 24.44 & 10.28 & Wan2.2~\cite{wan2025} \\ 
\bottomrule
\end{tabular}
}
\end{table*}

\subsection{Results Across Object Counts}
Due to space limitations, the main paper only reports the overall quantitative comparison on FlashBench.
Here, we present detailed evaluations under different numbers of controlled objects, covering cases from 1–5 to more than 5 foreground objects.
As shown in Table~\ref{tab:addition_comparions1} and Table~\ref{tab:addition_comparions2}, the ControlNet variant of FlashMotion consistently surpasses all competing methods across all metrics, outperforming both multi-step and few-step baselines in terms of visual quality and trajectory accuracy.
When using a ResNet-based trajectory adapter, FlashMotion also achieves better visual quality than the previous SOTA method MagicMotion~\cite{Li_2025_ICCV}, though it still falls slightly short in trajectory accuracy due to the limited parameter capacity.

\begin{table*}[h]
\caption{Quantitative comparison results on FlashBench for scenes containing 1, 2, and 3 controlled objects. The detailed evaluations show that FlashMotion with a ControlNet-based adapter consistently outperforms all competing methods across all metrics, while the ResNet-based adapter also delivers superior visual quality compared to prior work.}
\resizebox{\linewidth}{!}{
\label{tab:addition_comparions1}
\begin{tabular}{@{}lccccccccc@{}}
\toprule
\multirow{2}{*}{\textbf{Methods}}  
& \multicolumn{3}{c}{\textbf{Obj\_Num=1}} 
& \multicolumn{3}{c}{\textbf{Obj\_Num=2}} 
& \multicolumn{3}{c}{\textbf{Obj\_Num=3}} 
\\ \cmidrule(l){2-10}
& FID($\downarrow$) & FVD($\downarrow$) & M/B IoU($\uparrow$)
& FID($\downarrow$) & FVD($\downarrow$) & M/B IoU($\uparrow$)
& FID($\downarrow$) & FVD($\downarrow$) & M/B IoU($\uparrow$)
\\ \midrule
\multicolumn{10}{c}{\textbf{MultiSteps (50 Steps)}} \\ \midrule
MagicMotion~\cite{Li_2025_ICCV}& 53.62 & 741.91 & 67.93/83.46 & 59.37 & 697.50 & \underline{61.05/73.47} & 52.44 & 563.38 & \underline{66.13/72.92} \\
Wan2.2 (ResNet)~\cite{wan2025}    & 49.01 & 599.93 & 61.10/76.34 & 56.19 & 582.42 & 51.49/62.07 & 57.39 & 566.53 & 50.06/56.75 \\
Wan2.2 (ControlNet)~\cite{wan2025}& 50.04 & 594.54 & 66.07/83.98 & \underline{51.20} & 591.56 & 59.64/73.18 & 49.49 & 547.90 & 62.64/70.01 \\
DragAnything~\cite{wu2024draganything}       & 76.28 & 1076.20 & 62.70/74.88 & 91.08 & 1196.46 & 53.34/63.06 & 89.26 & 1099.45 & 54.01/57.55 \\
SG-I2V~\cite{namekata2024sgi2v}             & 70.20 & 984.94  & 64.09/76.45 & 78.93 & 926.79  & 47.16/57.04 & 73.08 & 891.52  & 48.31/54.25 \\
Tora~\cite{zhang2025tora}               & 73.15 & 902.55  & 58.24/69.00 & 80.27 & 939.72  & 46.45/57.47 & 82.54 & 869.43  & 46.80/52.66 \\
LeviTor~\cite{wang2025levitor}            &128.25 &1318.56  & 49.63/59.73 &127.24 &1124.07  & 38.09/44.82 &131.60 &1252.00  & 35.65/39.08 \\
\midrule
\multicolumn{10}{c}{\textbf{FewSteps (4 Steps) — Adapter: ResNet}} \\ \midrule
DMD~\cite{yin2024improved}                &64.71 &709.74 &55.34/74.30 &63.28 &687.09 &45.21/59.62 &64.03 &636.34 &43.08/53.14 \\
GAN~\cite{goodfellow2014generative}                &79.73 &728.35 &54.58/66.52 &77.25 &700.88 &41.38/51.34 &74.52 &673.58 &41.46/48.80 \\
LCM~\cite{luo2023latent}                &58.97 &875.26 &64.61/80.06 &72.26 &1032.56 &56.40/68.58 &65.52 &1033.12 &53.52/59.67 \\
\textbf{FlashMotion}
                   &\underline{46.64} &\underline{509.36} &\underline{68.02}/\textbf{84.86} &51.21 &\underline{497.62} &60.27/73.08 &\underline{44.41} &\underline{433.60} &63.40/71.59 \\
\midrule
\multicolumn{10}{c}{\textbf{FewSteps (4 Steps) — Adapter: ControlNet}} \\ \midrule
\rowcolor{gray!10} DMD~\cite{yin2024improved} / GAN~\cite{goodfellow2014generative} & \multicolumn{9}{c}{OOM} \\
LCM~\cite{luo2023latent}                &61.13 &851.48 &62.83/76.15 &76.41 &929.77 &56.68/66.79 &69.65 &831.79 &57.86/63.51 \\
\textbf{FlashMotion}
                   &\textbf{44.97} &\textbf{465.86} &\textbf{68.44}/\underline{84.51} &\textbf{46.16} &\textbf{437.18} &\textbf{63.87/76.99} &\textbf{42.20} &\textbf{422.16} &\textbf{66.45/73.91} \\ 
\bottomrule
\end{tabular}
}
\end{table*}
\begin{table*}[h]
\caption{Quantitative comparison results on FlashBench for scenes containing 4, 5, and above 5 controlled objects. The detailed evaluations show that FlashMotion with a ControlNet-based adapter consistently outperforms all competing methods across all metrics, while the ResNet-based adapter also delivers superior visual quality compared to prior work.}
\label{tab:addition_comparions2}
\resizebox{\linewidth}{!}{
\begin{tabular}{@{}lccccccccc@{}}
\toprule
\multirow{2}{*}{\textbf{Methods}}  
& \multicolumn{3}{c}{\textbf{Obj\_Num=4}} 
& \multicolumn{3}{c}{\textbf{Obj\_Num=5}} 
& \multicolumn{3}{c}{\textbf{Obj\_Num\textgreater5}} 
\\ \cmidrule(l){2-10}
& FID($\downarrow$) & FVD($\downarrow$) & M/B IoU($\uparrow$)
& FID($\downarrow$) & FVD($\downarrow$) & M/B IoU($\uparrow$)
& FID($\downarrow$) & FVD($\downarrow$) & M/B IoU($\uparrow$)
\\ \midrule
\multicolumn{10}{c}{\textbf{MultiSteps (50 Steps)}} \\ \midrule
MagicMotion~\cite{Li_2025_ICCV}        &45.67 &546.40 &\underline{70.29/73.21} &44.41 &450.10 &\underline{73.86/76.93} &44.41 &409.25 &\underline{69.29/62.35} \\
Wan2.2 (ResNet)~\cite{wan2025}    &61.69 &575.65 &50.89/53.93 &52.04 &476.04 &55.56/56.98 &41.59 &453.60 &44.31/41.03 \\
Wan2.2 (ControlNet)~\cite{wan2025}&49.25 &503.03 &66.15/68.65 &43.58 &409.57 &70.70/70.94 &37.11 &406.06 &67.27/61.05 \\
DragAnything~\cite{wu2024draganything}       &75.00 &997.03 &59.97/60.23 &83.35 &812.67 &62.92/61.48 &97.48 &1006.25 &56.95/49.51 \\
SG-I2V~\cite{namekata2024sgi2v}             &64.83 &861.49 &50.87/55.46 &65.91 &713.41 &54.21/55.83 &66.22 &828.14 &36.75/35.52 \\
Tora~\cite{zhang2025tora}               &65.25 &737.03 &46.28/51.05 &73.88 &714.65 &52.76/54.55 &93.60 &1073.05 &37.98/36.89 \\
LeviTor~\cite{wang2025levitor}            &167.97 &1774.66 &35.10/34.02 &185.75 &2015.57 &33.33/30.34 &135.75 &1287.67 &24.23/23.41 \\
\midrule
\multicolumn{10}{c}{\textbf{FewSteps (4 Steps) — Adapter: ResNet}} \\ \midrule
DMD~\cite{yin2024improved}                &66.08 &749.99 &41.38/48.44 &67.03 &697.32 &42.02/47.94 &52.62 &671.65 &32.74/32.74 \\
GAN~\cite{goodfellow2014generative}                &69.55 &571.76 &45.87/50.43 &65.83 &500.22 &48.67/52.49 &59.83 &584.86 &31.00/30.78 \\
LCM~\cite{luo2023latent}                &62.24 &959.97 &57.30/57.89 &58.71 &869.45 &56.78/58.98 &49.66 &780.45 &43.51/40.21 \\
\textbf{FlashMotion}
                   &\underline{38.47} &\underline{411.71} &66.58/67.87 &\underline{39.53} &\underline{326.98} &68.67/79.78 &\underline{37.07} &\underline{384.06} &56.92/52.02 \\
\midrule
\multicolumn{10}{c}{\textbf{FewSteps (4 Steps) — Adapter: ControlNet}} \\ \midrule
\rowcolor{gray!10} DMD~\cite{yin2024improved} / GAN~\cite{goodfellow2014generative} & \multicolumn{9}{c}{OOM} \\
LCM~\cite{luo2023latent}                &60.28 &752.48 &63.18/63.38 &56.29 &637.06 &66.48/65.66 &53.64 &541.55 &60.79/53.62 \\
\textbf{FlashMotion}
                   &\textbf{36.62} &\textbf{367.24} &\textbf{71.81/75.49} &\textbf{35.19} &\textbf{294.47} &\textbf{74.94/76.98} &\textbf{32.72} &\textbf{305.68} &\textbf{69.43/64.50} \\
\bottomrule
\end{tabular}
}
\end{table*}

\subsection{More Qualitative Results}
\begin{figure*}[ht]
    \centering
    \includegraphics[width=\linewidth]{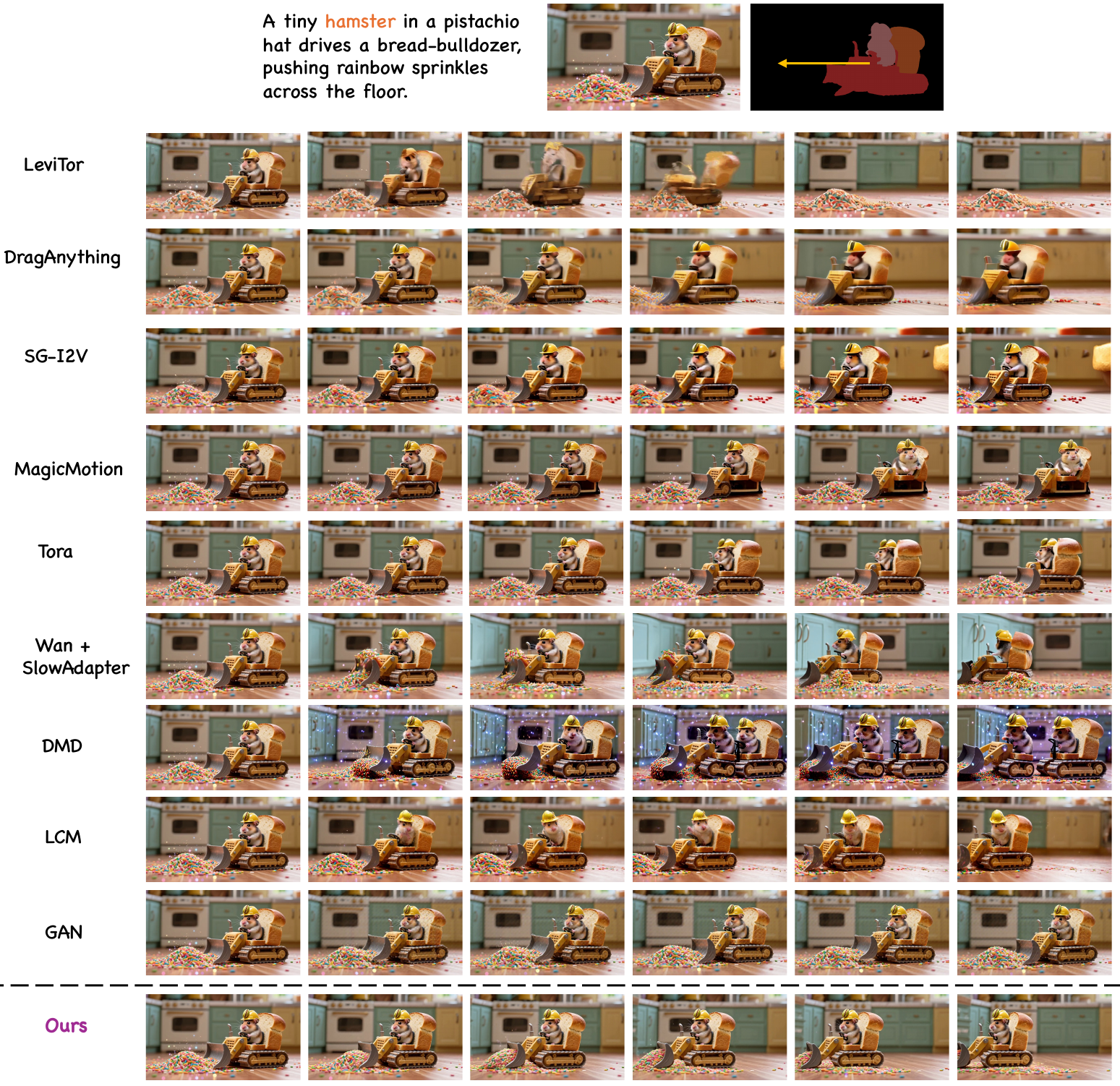}
     \caption{Qualitative Comparisons results with different methods.}
    \label{fig:Comparions_supp1}
\end{figure*}

\begin{figure*}[ht]
    \centering
    \includegraphics[width=\linewidth]{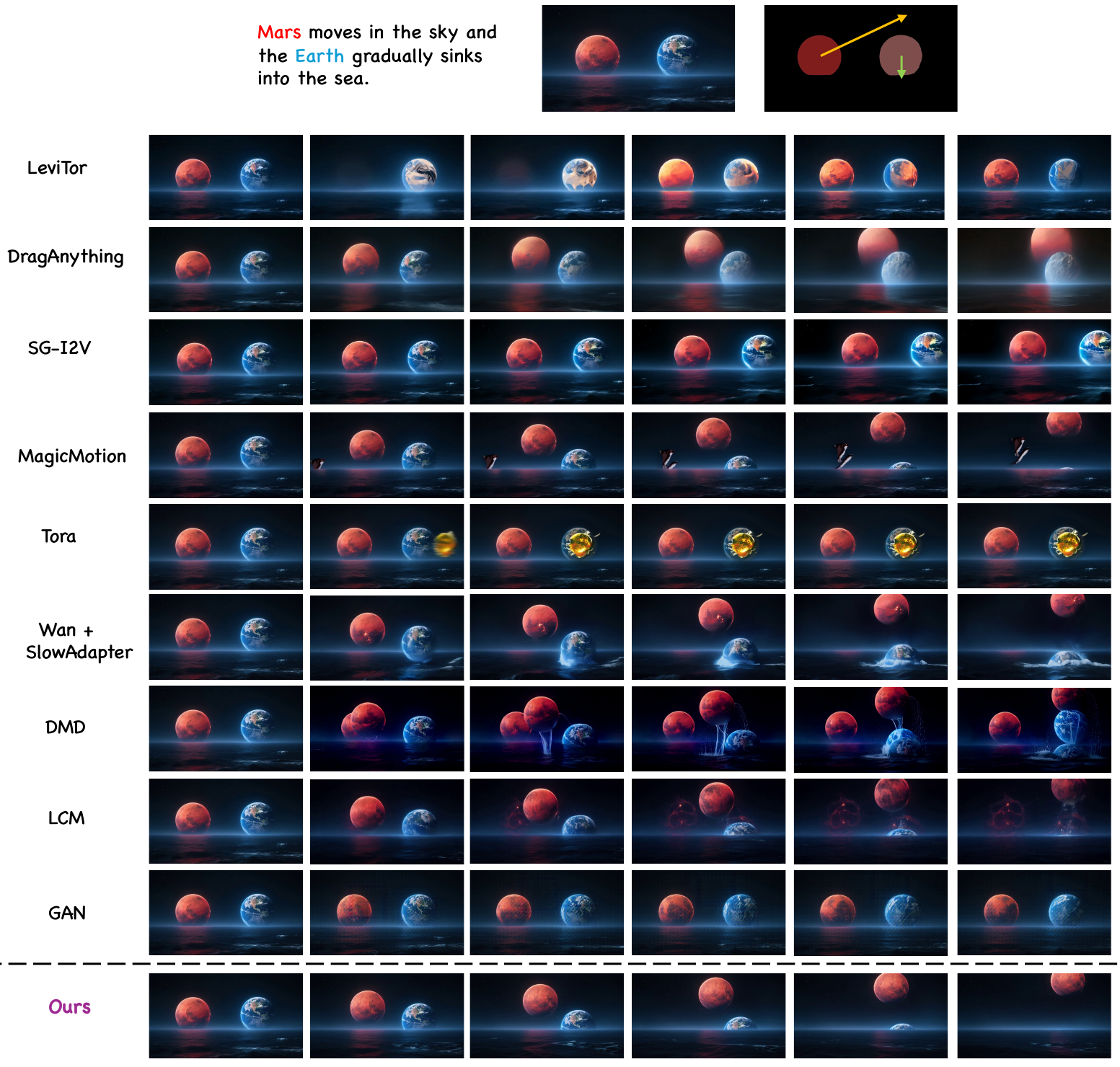}
     \caption{Qualitative Comparisons results with different methods.}
    \label{fig:Comparions_supp2}
\end{figure*}

\begin{figure*}[ht]
    \centering
    \includegraphics[width=\linewidth]{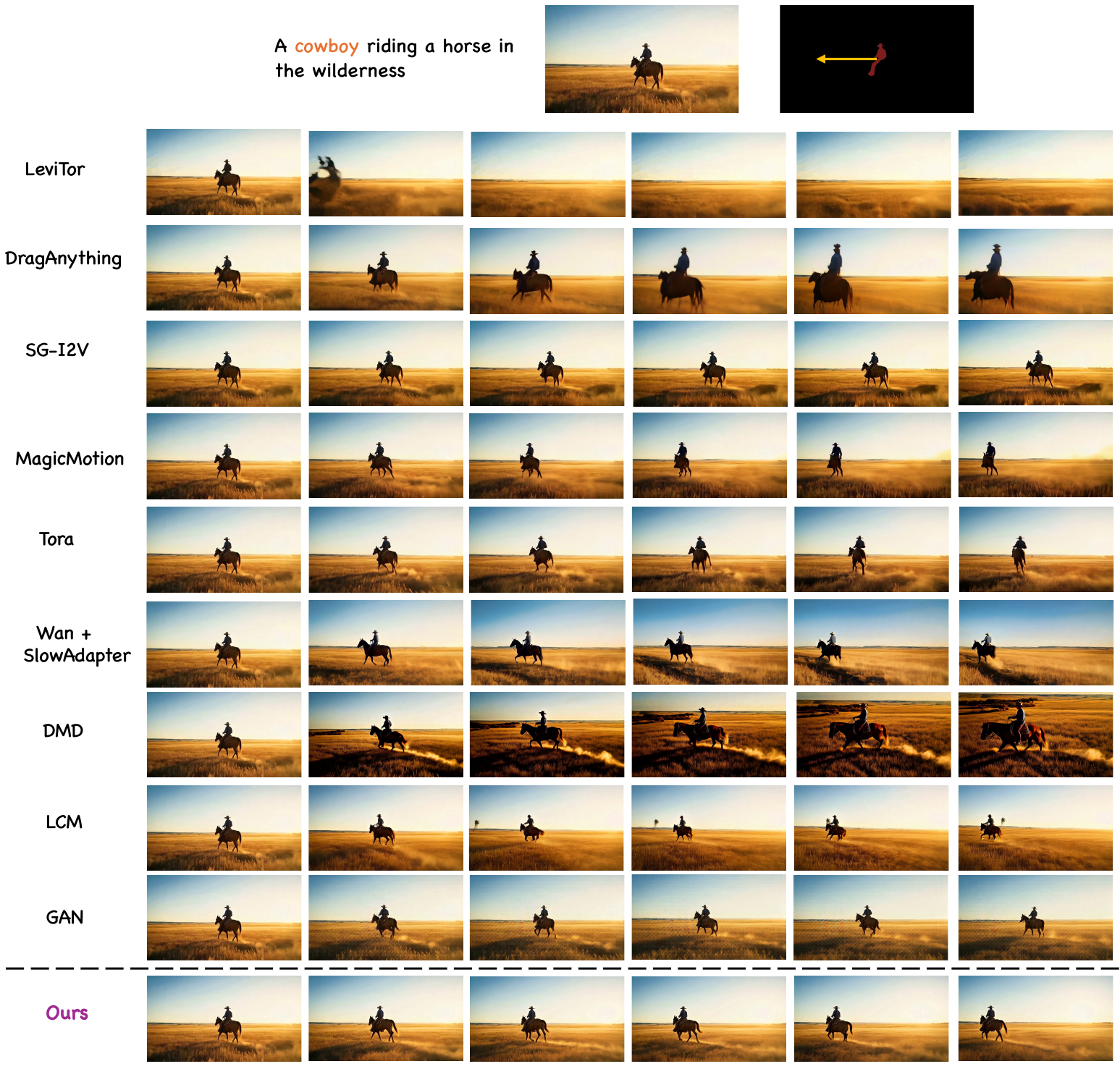}
     \caption{Qualitative Comparisons results with different methods.}
    \label{fig:Comparions_supp3}
\end{figure*}

\begin{figure*}[ht]
    \centering
    \includegraphics[width=\linewidth]{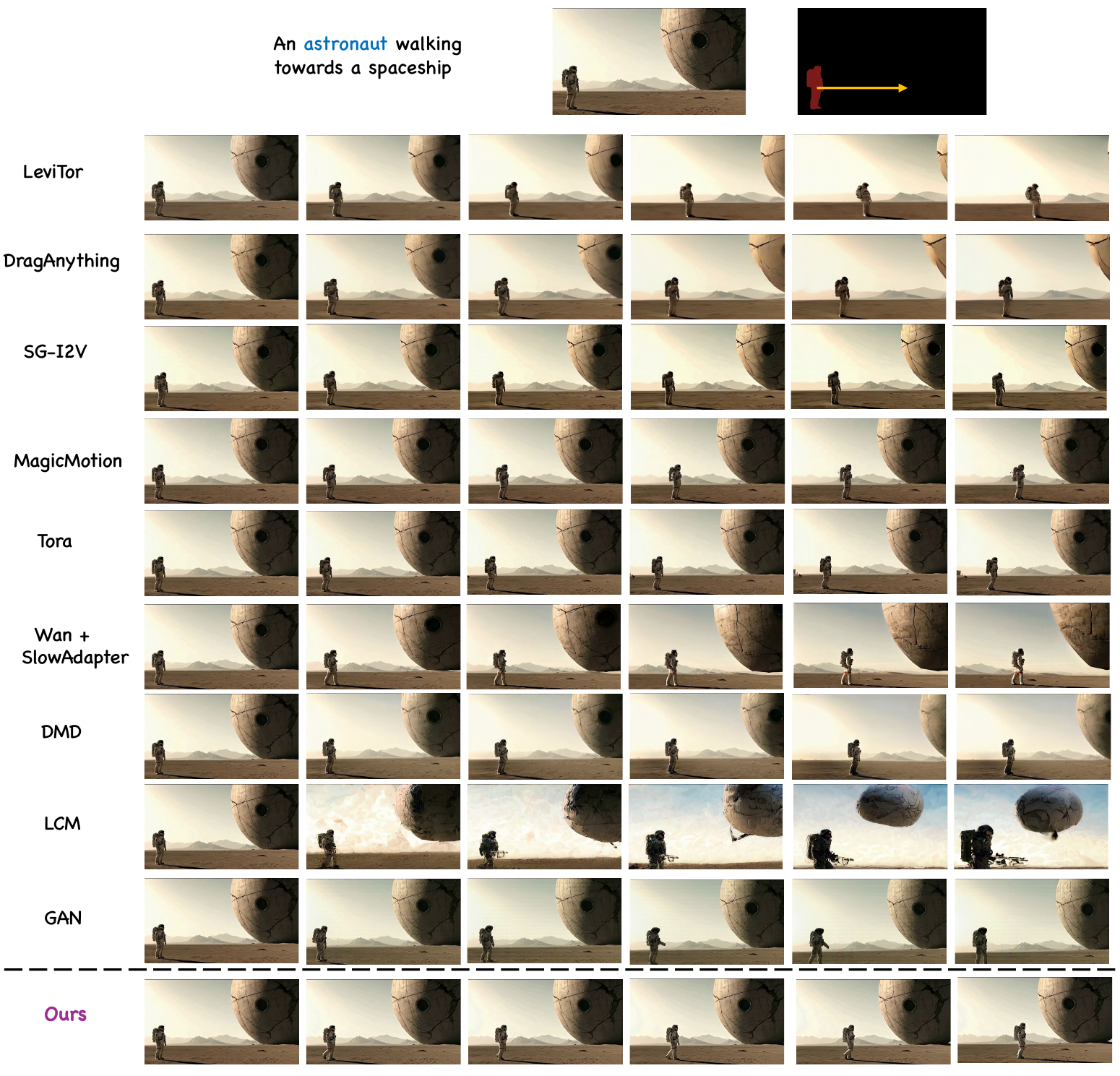}
     \caption{Qualitative Comparisons results with different methods.}
    \label{fig:Comparions_supp4}
\end{figure*}

\begin{figure*}[ht]
    \centering
    \includegraphics[width=\linewidth]{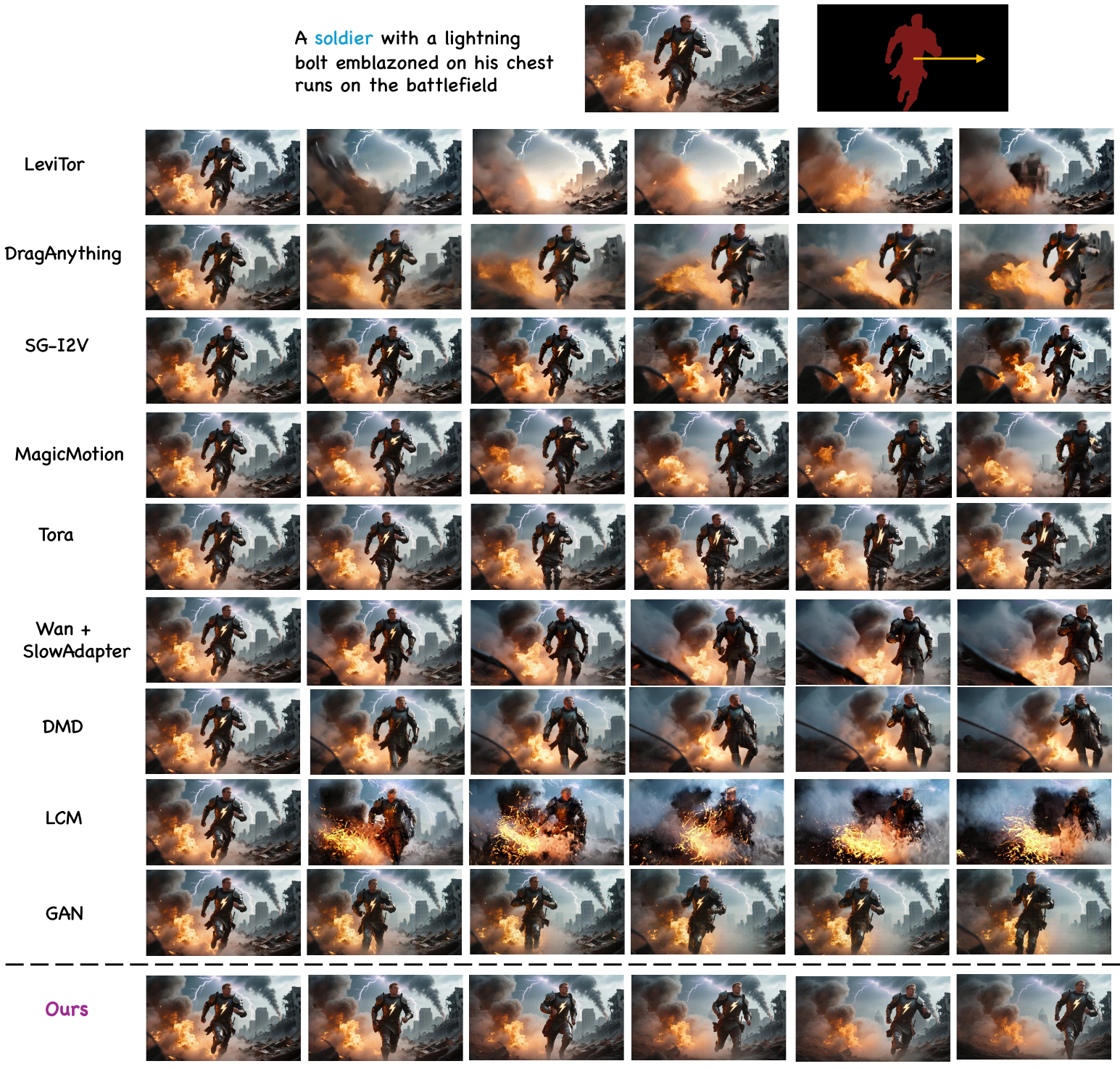}
     \caption{Qualitative Comparisons results with different methods.}
    \label{fig:Comparions_supp5}
\end{figure*}

\begin{figure*}[ht]
    \centering
    \includegraphics[width=\linewidth]{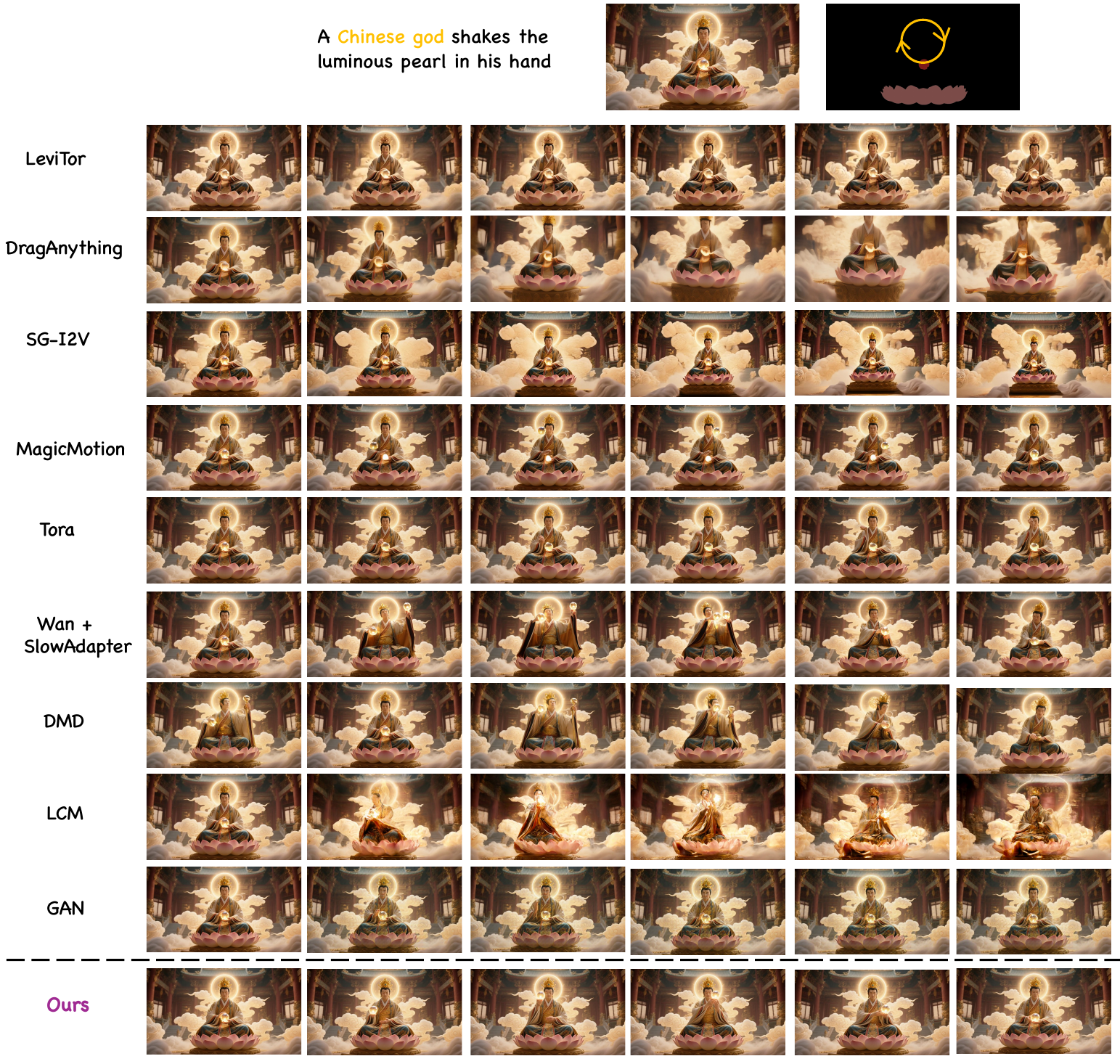}
     \caption{Qualitative Comparisons results with different methods.}
    \label{fig:Comparions_supp6}
\end{figure*}

\begin{figure*}[ht]
    \centering
    \includegraphics[width=\linewidth]{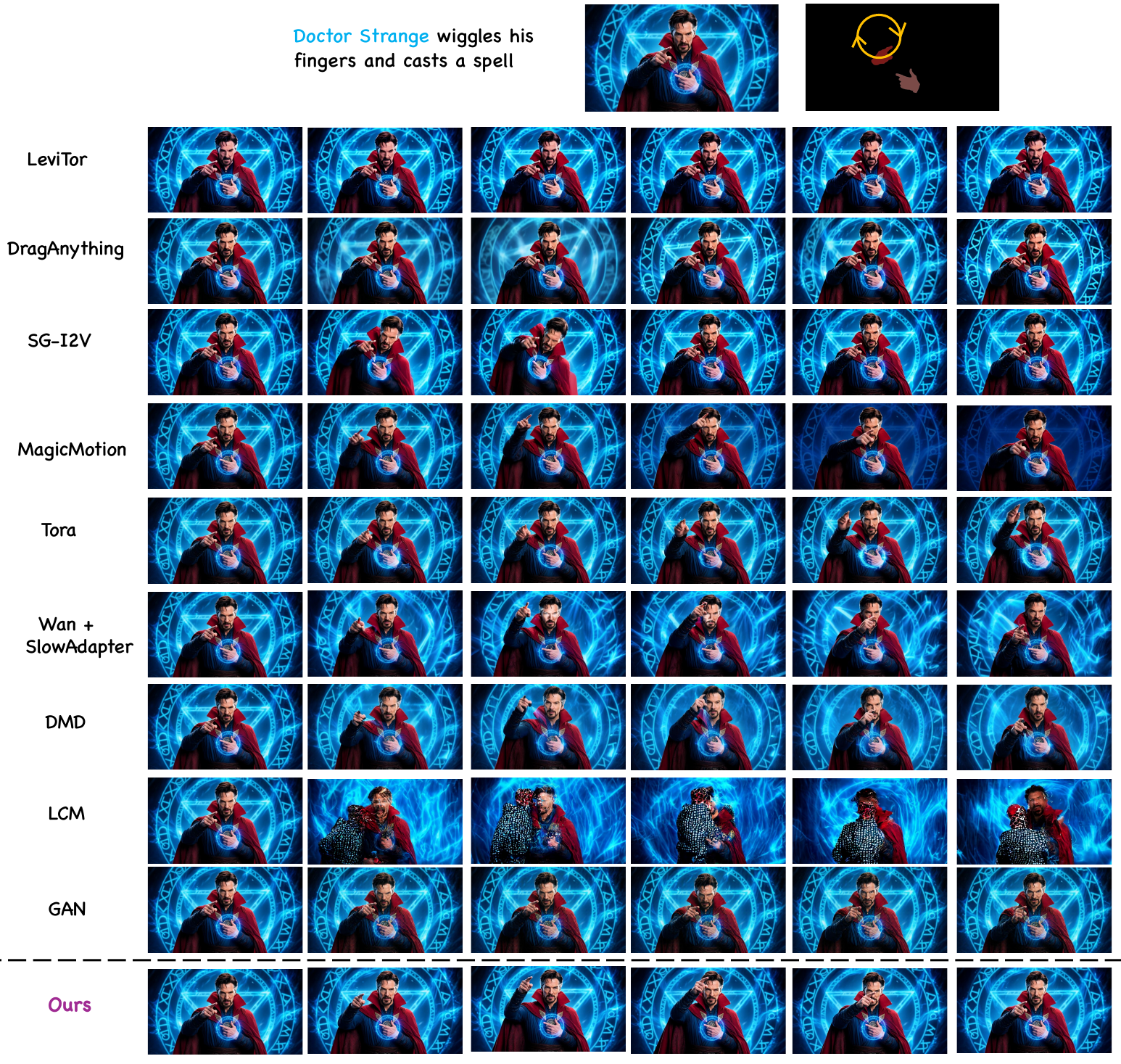}
     \caption{Qualitative Comparisons results with different methods.}
    \label{fig:Comparions_supp7}
\end{figure*}

In this section, we present additional qualitative comparisons with previous methods. As illustrated in Figs.~\ref{fig:Comparions_supp1}–\ref{fig:Comparions_supp7}, FlashMotion accurately controls object trajectories and produces high-quality videos, whereas the other approaches exhibit notable artifacts and inconsistencies. For full video results, please refer to “Supplementary video.mp4” in the supplementary material.
\section{Case Studies}
\subsection{Different Styles}
\begin{figure*}[ht]
    \centering
    \includegraphics[width=\linewidth]{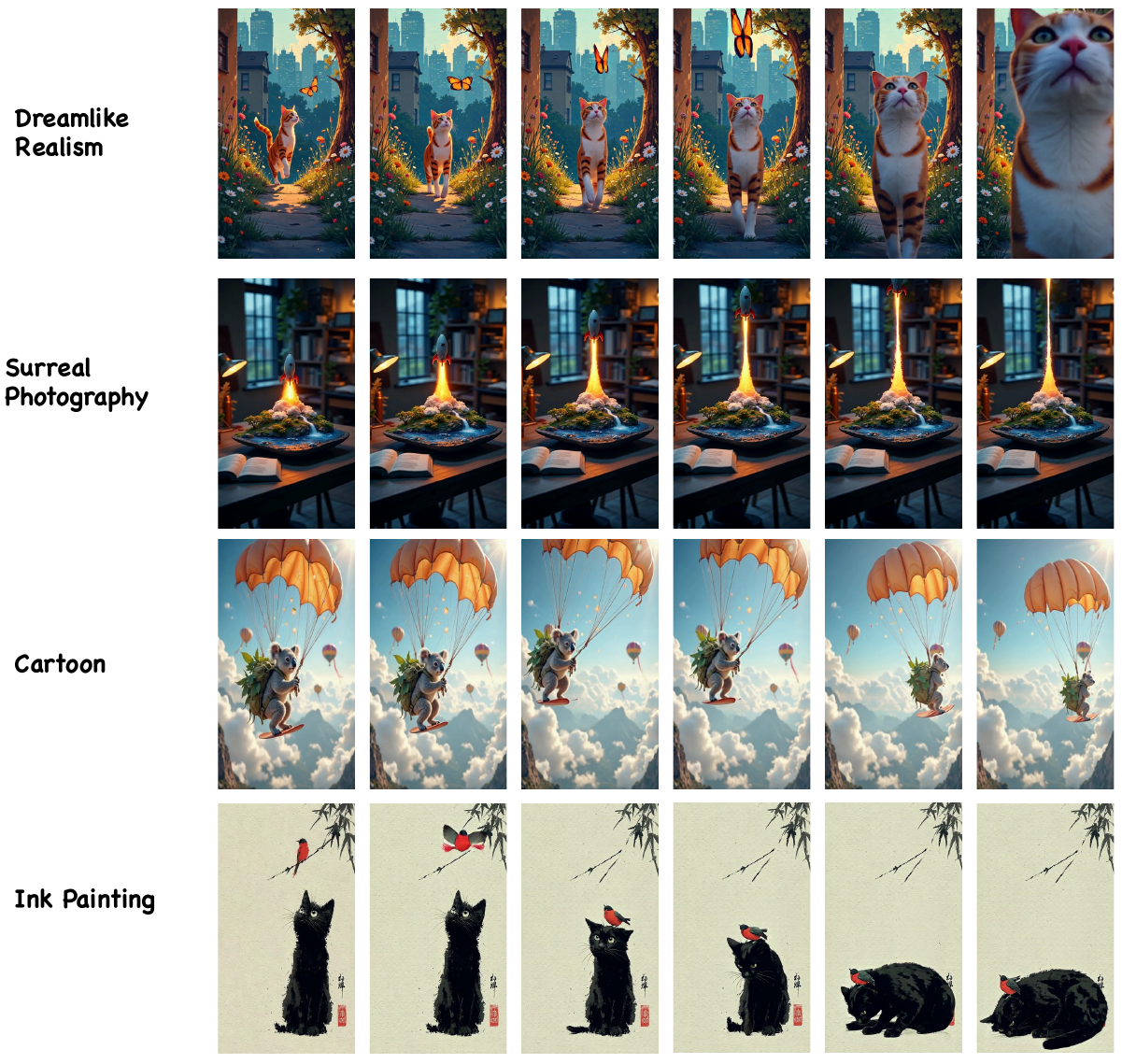}
     \caption{FlashMotion supports generating videos of different styles.}
    \label{fig:style}
\end{figure*}
As shown in Fig.~\ref{fig:style}, FlashMotion supports generating videos across diverse visual styles, including dreamlike realism, surreal miniature photography, 3D cartoon rendering, and Eastern ink-wash painting.
To better demonstrate the model’s robustness and its ability to maintain consistent motion across challenging layouts, we deliberately choose vertically oriented images instead of horizontal ones. These examples collectively illustrate FlashMotion’s strong adaptability to various artistic domains while preserving coherent structure and motion.
\subsection{Camera Control}
\begin{figure*}[ht]
    \centering
    \includegraphics[width=\linewidth]{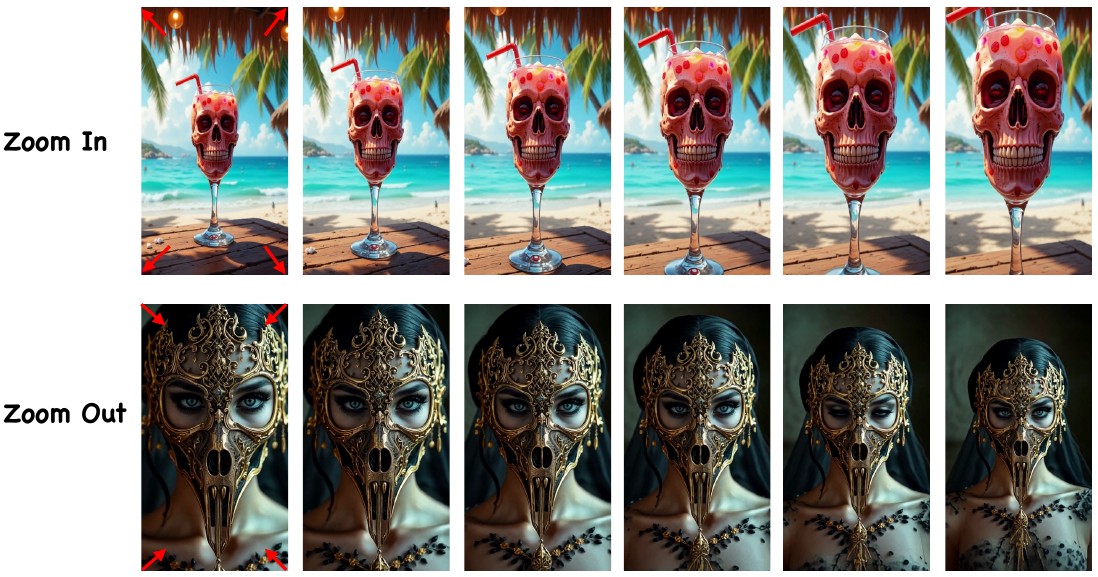}
     \caption{FlashMotion enables controllable camera movements, such as zooming in or out, by adjusting the bounding box size of the foreground object (e.g., the cup or the woman’s mask).}
    \label{fig:camera1}
\end{figure*}
\begin{figure*}[ht]
    \centering
    \includegraphics[width=\linewidth]{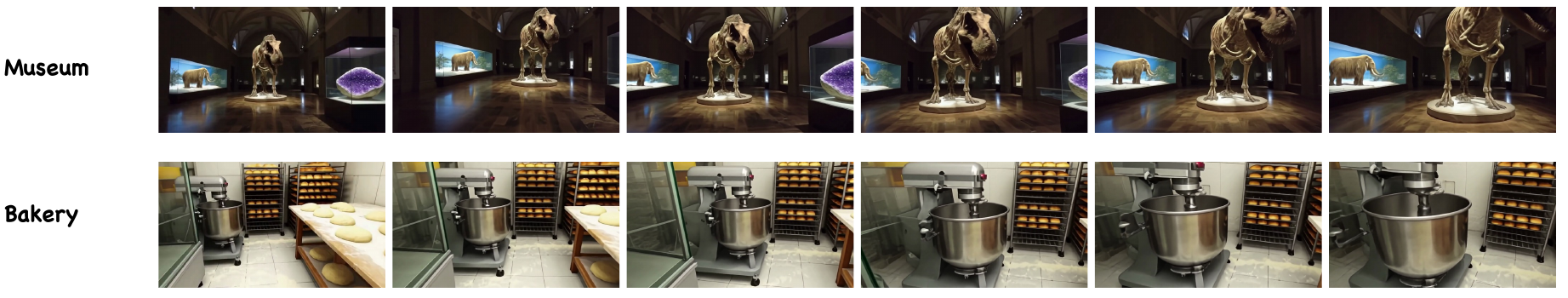}
     \caption{FlashMotion supports scene navigation in various environments—such as a bakery or a museum—by manipulating the bounding boxes of key objects, including the dinosaur, the mammoth, and the industrial mixer.}
    \label{fig:camera2}
\end{figure*}
FlashMotion supports camera control operations such as zooming in and zooming out. As shown in Fig.\ref{fig:camera1}, the camera motion can be adjusted by manipulating the bounding box size of the foreground object, such as the cup or the woman’s mask.
Furthermore, as illustrated in Fig.\ref{fig:camera2}, users can navigate scenes—like a bakery or a museum—by controlling the bounding boxes of objects such as the dinosaur, the mammoth, or the industrial mixer.

\section{More Details on FlashBench}
\begin{figure*}[ht]
    \centering
    \includegraphics[width=\linewidth]{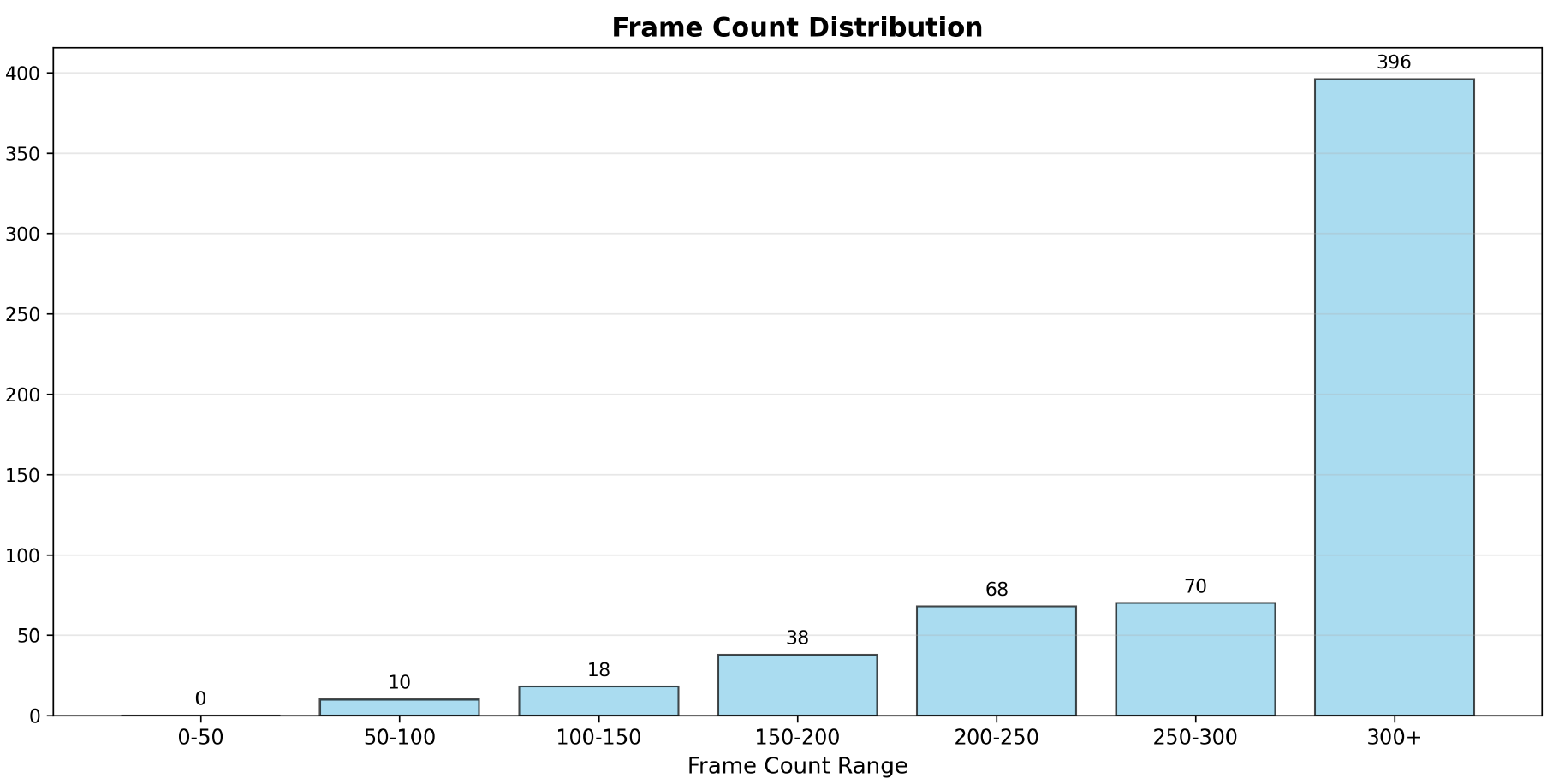}
    \caption{Distribution of video frame counts in FlashBench, demonstrating its support for evaluating long video generation.}
    \label{fig:benchmark}
\end{figure*}

FlashBench comprises 600 videos, grouped into six categories based on the number of foreground objects (ranging from 1–5 and more than 5). To offer a more comprehensive analysis of the dataset, we further visualize the distributions of video lengths as shown in Fig.~\ref{fig:benchmark}, demonstrating its support for evaluating long video generation.
\end{document}